\documentclass[10pt,twocolumn,letterpaper]{article}

\usepackage{titlesec}

\makeatletter
\@addtoreset{section}{part}
\makeatother
\titleformat{\part}[display]
{\normalfont\LARGE\bfseries\centering}{}{0pt}{}

\usepackage[pagenumbers]{cvpr} 

\usepackage{times}
\usepackage{epsfig}
\usepackage{graphicx}
\usepackage{amsmath}
\usepackage{amssymb}
\usepackage{multirow}
\usepackage{booktabs}
\usepackage{rotating}
\usepackage{caption}
\usepackage{overpic}
\newcommand{\tabincell}[2]{
\begin{tabular}{@{}#1@{}}#2\end{tabular}
}

\usepackage{color, colortbl}
\def \bbb{\color{blue}}
\def \rrr{\color{red}}
\def \ggg{\color{green}}
\def \kkk{\color{black}}
\definecolor{Gray}{gray}{0.9}
\definecolor{mygray}{gray}{.9}

\newcommand\blfootnote[1]{%
\begingroup
\renewcommand\thefootnote{}\footnote{#1}%
\addtocounter{footnote}{-1}%
\endgroup
}

\newcommand{\figref}[1]{Fig.\ref{#1}}
\newcommand{\tabref}[1]{Tab.\ref{#1}}
\newcommand{\secref}[1]{$\S$\ref{#1}}
\newcommand{\supp}[1]{\textcolor{magenta}{#1}}
\def\ourmodel{IS-Net}
\def \ourdataset{DIS5K}

\usepackage{algorithmic}
\usepackage[linesnumbered,boxed]{algorithm2e}

\usepackage[pagebackref=true,breaklinks=true,letterpaper=true,colorlinks=true,bookmarks=false]{hyperref}

\usepackage{url} 

\def\cvprPaperID{4674} 
\def\confName{CVPR}
\def\confYear{2022}

\begin{document}

\title{Highly Accurate Dichotomous Image Segmentation}










\author{Xuebin Qin\\
MBZUAI\\
Abu Dhabi, UAE\\
{\tt\small xuebinua@gmail.com}
\and
Hang Dai\\
MBZUAI\\
Abu Dhabi, UAE\\
{\tt\small hang.dai@mbzuai.ac.ae}
\and
Xiaobin Hu\\
TUM\\
Munich, Germany\\
{\tt\small xiaobin.hu@tum.de}
\and
Deng-Ping Fan~$*$\\
ETH Zurich,\\
Switzerland\\
{\tt\small dengpfan@gmail.com}
\and
Ling Shao\\
Terminus Group\\
China\\
{\tt\small ling.shao@ieee.org}
\and
Luc Van Gool\\
ETH Zurich, \\
Switzerland\\
{\tt\small vangool@vision.ee.ethz.ch}
}

\twocolumn[{%
\renewcommand\twocolumn[1][]{#1}%
\maketitle
\vspace{-35pt}
\begin{center}
    \centering
    \captionsetup{type=figure}
    \includegraphics[width=\textwidth,height=.209\textwidth]{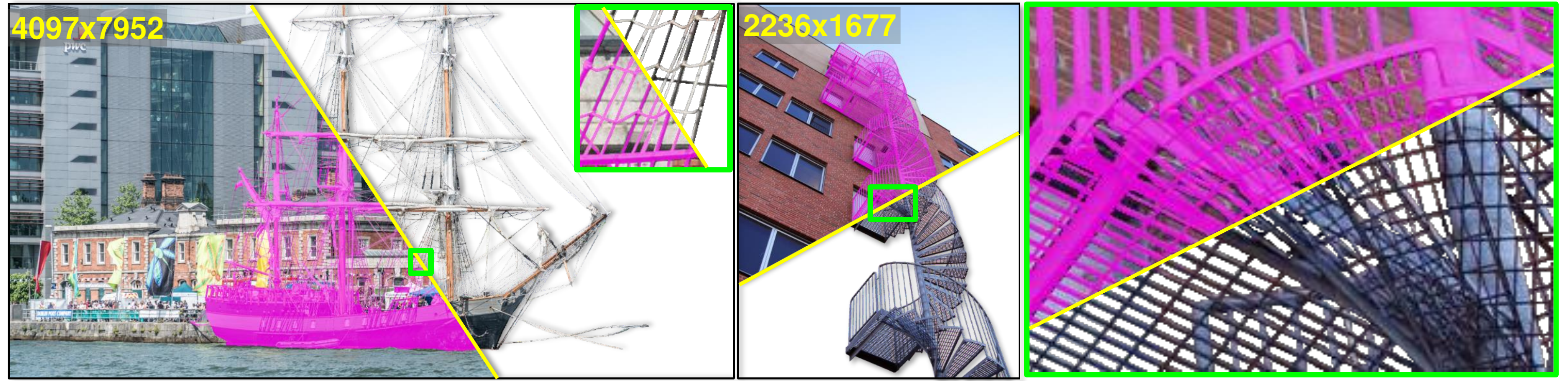}
    \captionof{figure}{\small Sample images (backgrounds partially removed by ground truth (GT) masks) from our \ourdataset~dataset. Zoom-in for best view.
    }
    \label{fig:intro}
\end{center}%
}]


\begin{abstract}

We present a systematic study on a new task 
called dichotomous image segmentation (DIS), 
which aims to segment highly accurate objects from natural images. 
To this end, we collected the first large-scale dataset, called \textbf{\ourdataset}\blfootnote{* Corresponding author.}, 
which contains 5,470 high-resolution (\eg, 2K, 4K or larger) images 
covering \textit{camouflaged}, \textit{salient}, or \textit{meticulous objects} in various backgrounds. All images are annotated with extremely fine-grained labels. 
In addition, we introduce a simple intermediate supervision 
baseline (\textbf{\ourmodel}) using both 
feature-level and mask-level guidance for DIS model training. 
Without tricks, \ourmodel~outperforms various 
cutting-edge baselines on the proposed \ourdataset, 
making it a general self-learned supervision network 
that can help facilitate future research in DIS.
Further, we design a new metric called human correction efforts (\textbf{HCE}) which 
approximates the number of mouse clicking operations required to correct the false 
positives and false negatives. HCE is utilized to measure the gap between models and real-world applications and thus can complement existing metrics.
Finally, we conduct the largest-scale benchmark, 
evaluating 16 representative segmentation models, 
providing a more insightful discussion regarding object complexities,
and showing several potential applications (\eg, background removal, art design, 3D reconstruction).
Hoping these efforts can open up promising directions for both academic and industries.
Our \ourdataset~dataset, \ourmodel~baseline, HCE metric, and the complete benchmarks will be made publicly available at: \url{https://xuebinqin.github.io/dis/index.html}.

\end{abstract}

\section{Introduction}
In many years, the accuracy of annotations in computer vision datasets that drive a tremendous amount of Artificial Intelligence (AI) models satisfy the requirements of machine perceiving systems to some extent. However, AI has entered an era of demanding highly accurate outputs from computer vision algorithms to support delicate human-machine interaction and immersed virtual life.
Image segmentation, as one of the most fundamental techniques in computer vision, plays a vital role in enabling the machines to perceive and understand the real world. Compared with image classification \cite{imagenet_cvpr09,DBLP:conf/nips/KrizhevskySH12,simonyan2014very} and object detection \cite{DBLP:conf/cvpr/GirshickDDM14,girshick2015fast,ren2015faster}, it can provide more geometrically accurate descriptions of the targets used in a wide range of applications, such as image editing \cite{goferman2012context}, 3D reconstruction \cite{Liu_2021_CVPR}, augmented reality (AR) \cite{qin2021boundary}, satellite image analysis \cite{DBLP:journals/remotesensing/WeiLLZCJZY21}, medical image processing \cite{ronneberger2015u}, robot manipulation \cite{DBLP:conf/rss/Chen0LH21}, \etc. 
We can categorize the above applications as ``light’’ (\eg, image editing and image analysis) and ``heavy’’ (\eg, manufacturing and surgical robots), based on their immediate affects on real-world objects. 
The ``light’’ applications (\figref{fig:intro}) are relatively tolerant to the segmentation deflects and failures because these issues mainly lead to more labors and time costs, which are usually affordable.
While, in the ``heavy’’ applications, those deflects or failures are more likely to cause serious consequences, which are usually physic damages on objects or injuries, sometimes fatal for creatures, \eg, humans and animals. Hence, these applications require the models to be \textit{highly accurate} and \textit{robust}. 
Currently, most of the segmentation models are still less applicable in those ``heavy'' applications because of the accuracy and robustness issues, which restricts the segmentation techniques from playing more essential roles in broader applications. 
Here, \textbf{our goal} is to address the ``heavy'' and ``light'' applications in a general framework, we called this task as \textit{dichotomous image segmentation (DIS)}, which aims to segment highly accurate objects from the nature images.

\begin{figure*}[t!]
    \centering
    \includegraphics[width=\linewidth]{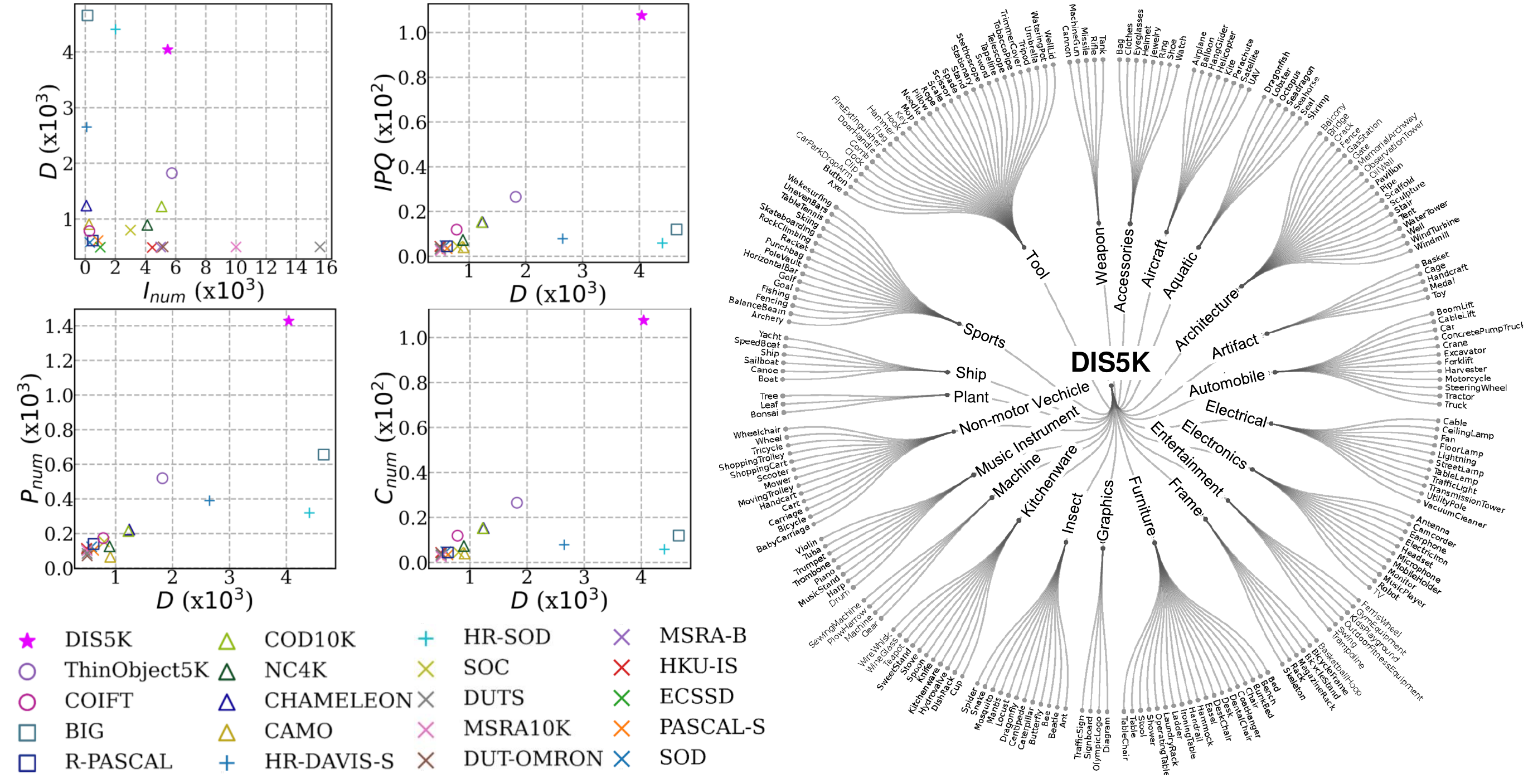}
    \vspace{-20pt}
    \caption{\small \textbf{Left}: Correlations between different complexities. \textbf{Right}: Categories and groups of our DIS5K dataset. Zoom-in for better view. Please refer to \secref{sec:DataCol} for details.}
    \label{fig:DIS-cats-grps}
    \vspace{-10pt}
\end{figure*}

However, existing image segmentation tasks mainly focus on segmenting objects 
with specific characteristics, \eg, salient \cite{wang2017learning,zeng2019towards,HRRN_ICCV2021}, camouflaged \cite{le2019anabranch,chameleon,fan2020camouflaged}, meticulous \cite{liew2021deep,yang2020meticulous} or specific categories \cite{ronneberger2015u,MnihThesis,shen2016automatic,MODNet,DBLP:journals/corr/abs-2108-11515}. Most of them have the same input/output formats, and barely use exclusive mechanisms designed for segmenting targets in their models, which means almost all tasks are dataset-dependent. 
Thus, we propose to formulate \textbf{a category-agnostic DIS task defined on non-conflicting annotations for accurately segmenting objects with different structure complexities, regardless of their characteristics}. 
Compared with semantic segmentation~\cite{everingham2010pascal,lin2014microsoft,Cordts2016Cityscapes,zhou2017scene,qi2021open}, the proposed DIS task usually focuses on images with single or a few targets, from which getting richer accurate details of each target is more feasible. To this end, we provide four \textbf{contributions}:
\begin{enumerate}
    \vspace{-3pt}
    \item 
    A large-scale, extendable DIS dataset, \textbf{DIS5K}, contains 5,470 high-resolution images paired with highly accurate binary segmentation masks.
    \vspace{-4pt}
    \item 
    A novel baseline \textbf{\ourmodel} built with our intermediate supervision reduces over-fitting by enforcing direct feature synchronization in high-dimensional feature spaces.
    \vspace{-4pt}
    \item 
    A newly designed human correction efforts (\textbf{HCE}) metric measures the barriers between model predictions and real-world applications by counting the human interventions needed to correct the faulty regions.
    \vspace{-4pt}
    \item Based on the new DIS5K, we establish the complete DIS \textbf{benchmark}, making ours the most extensive DIS investigation. We compared our \ourmodel~with 16 cutting-edge segmentation models and showed promising results for background removal and 3D reconstruction applications.
\end{enumerate}

\section{Related Work}\label{sec:RelatedWork}
\noindent\textbf{Tasks and Datasets} of image segmentation are closely related in deep learning era. Some of the segmentation tasks like~\cite{DBLP:journals/pami/ChengMHTH15,wang2017learning,fan2021concealed,liew2021deep,yang2020meticulous,shen2016automatic,DBLP:journals/corr/abs-2108-11515,MnihThesis}, are even directly built upon the datasets. Their problem formulations are exactly the same: $P = F(\theta,I)$, where $I$ and $P$ are the input image and the binary map output, respectively. 
However, the relevance between most of these tasks are rarely studied, which somehow restricts the models trained for certain tasks from being generalized to wider applications. Besides, the datasets used in different tasks are not exclusive, 
which shows a unified task for \textit{dichotomous image segmentation} (DIS) is possible. 
However, most of the existing datasets are built on low-resolution images with objects of simple structures. There lacks an dataset built on the accurately labeled high-resolution images which contain objects with diversified shape complexities from different categories. 

\noindent\textbf{Models} are often struggling with the conflicts between stronger representative capabilities and higher risks of over-fitting. To obtain more representative features, FCN-based models \cite{long2015fully}, Encoder-Decoder \cite{badrinarayanan2017segnet,ronneberger2015u}, Coarse-to-Fine \cite{wang2018detect}, Predict-Refine \cite{qin2019basnet,HRRN_ICCV2021}, Vision Transformer \cite{zheng2020rethinking} and so on are developed. Besides, many real-time models are designed~\cite{zhao2018icnet,yu2018bisenet,li2019dfanet,orsic2019defense,hu2020temporally,Fan_2021_CVPR,nirkin2020hyperseg} to balance the performance and the time costs. 
Other methods, such as weights regularization \cite{Goodfellow-et-al-2016}, dropout \cite{DBLP:journals/jmlr/SrivastavaHKSS14}, dense supervision \cite{lee2015deeply,qin2020u2,xie2015holistically}, and hybrid loss \cite{luc2016semantic,qin2019basnet,zhao2019egnet}, focus on alleviating the over-fitting. Dense supervision is one of the most effective ways for reducing the over-fitting. However, supervising the side outputs from the intermediate deep features may not be the best option because the supervision is weakened by the conversion from deep features (multi-channel) to side outputs (one-channel).

\noindent\textbf{Evaluation Metrics} can be categorized as \textit{region-based} (\eg, IoU or Jaccard index \cite{IoUJaccard}, F-measure \cite{DBLP:conf/muc/Chinchor92,Rijsbergen1979} or Dice's coefficient \cite{Thorvald1948}, weighted F-measure \cite{Margolin2014HowTE}), \textit{boundary-based} (\eg, CM \cite{movahedi2010design}, boundary F-measure~\cite{DBLP:journals/pami/MartinFM04,perazzi2016benchmark,ehrig2005relaxed,mnih2010learning,saito2016multiple,zhang2018road,qin2019basnet}, boundary IoU \cite{cheng2021boundary}, boundary displacement error (BDE) \cite{DBLP:conf/eccv/FreixenetMRMC02}, Hausdorff distances \cite{Hausdorff1914,Henry1920,DBLP:conf/ifip7/BirsanT05}), \textit{structure-based} (\eg, S-measure \cite{fan2017structure}, E-measure \cite{fan2018enhanced,fan2021cognitive}), \textit{confidence-based} (\eg, MAE \cite{perazzi2012saliency}), \etc. They mainly measure the consistencies between the predictions and the ground truth from mathematical or cognitive perspectives. But the costs of synchronizing the predictions against the requirements in real-world applications are not well studied. 

\section{Proposed \ourdataset~Dataset}
\subsection{Data Collection and Annotation}\label{sec:DataCol}
\textbf{Data Collection.} To address the dataset issue (see \secref{sec:RelatedWork}), we build a highly accurate DIS dataset named \textbf{\ourdataset}. We first manually collected over 12,000 images from Flickr\footnote{Images with the license of ``Commercial use \& mods allowed''} based on our pre-designed keywords\footnote{Since the long-term goal of this research is to facilitate the ``safe'' and ``efficient'' interaction between the machines and our living/working environments, these keywords are mainly related to the common targets (\eg, bicycle, chair, bag, cable, tree, \etc) in our daily lives.}. 
Then, according to the structural complexities of the objects, 
we obtained 5,470 images covering 225 categories (\figref{fig:DIS-cats-grps}) 
in 22 groups. Note that the adopted selection strategy is similar to Zhou \etal~\cite{zhou2017places}. Most selected images only contain single objects to obtain rich and highly accurate structures and details. Meanwhile, the segmentation and labeling confusions caused by the co-occurrence of multiple objects from different categories are avoided to the greatest extent. Specifically, the image selection criteria can be summarized as follows: 
\begin{itemize}
    \item[$\bullet$] Cover more categories while reducing the number of ``redundant'' samples with simple structures, which other existing datasets have already covered.
    
    \item[$\bullet$] Enlarge the intra-category dissimilarities (See $\S$2.3 of the \supp{supplementary (SM)}) of the selected categories by adding more diversified intra-category images 
    (\figref{fig:datasets-qual}-f). 
    
    \item[$\bullet$] Include more categories with complicated structures, \eg, \textit{fence, stairs, cable, bonsai, tree,} \etc, which are common in our lives but not well-labeled 
    (\figref{fig:datasets-qual}-a) or neglected by other datasets due to labeling difficulties (\figref{fig:datasets-cmplx}). 
\end{itemize}

Therefore, the labeled targets in our \ourdataset~are mainly the ``\textit{foreground objects of the images defined by the pre-designed keywords}'' regardless of their characteristics \eg, \textit{salient, common, camouflaged, meticulous,} \etc. 

\begin{table*}[t!]
    \centering
    \scriptsize
    \caption{\small
    Data analysis of existing datasets. See \secref{sec:DA} for details.
    }\label{tab:my_label}
    \renewcommand{\arraystretch}{1.0}
    \setlength\tabcolsep{4.0pt}
    \resizebox{\textwidth}{!}{
    \begin{tabular}{r|r||r|rrr|rrr}
        \toprule
        \multirow{2}{*}{Task} & \multirow{2}{*}{Dataset} & Number & \multicolumn{3}{c|}{Image Dimension} & \multicolumn{3}{c}{Object Complexity}\\
        \cmidrule(lr){3-3}
        \cmidrule(lr){4-6} \cmidrule(lr){7-9}
         &  & $I_{num}$ & $H\pm\sigma_H$ & $W\pm\sigma_W$ & $D\pm\sigma_D$ & $IPQ\pm\sigma_{IPQ}$ & $C_{num}\pm\sigma_C$ & $P_{num}\pm\sigma_P$\\
        \hline 
        \multirow{12}{*}{SOD}  & SOD\cite{movahedi2010design}  &  300  &  366.87 $\pm$ 72.35  &  435.13 $\pm$ 72.35  &  578.28 $\pm$ 0.00  &  4.74 $\pm$ 3.89  &  2.25 $\pm$ 1.76  &  122.79 $\pm$ 62.97  \\
        & PASCAL-S\cite{li2014secrets}  &  850  &  387.63 $\pm$ 64.65  &  467.82 $\pm$ 61.46  &  613.22 $\pm$ 32.00  &  3.39 $\pm$ 2.46  &  5.14 $\pm$ 11.72  &  102.76 $\pm$ 70.09  \\
        & ECSSD\cite{yan2013hierarchical}  &  1000  &  311.11 $\pm$ 56.27  &  375.45 $\pm$ 47.70  &  492.75 $\pm$ 19.78  &  3.26 $\pm$ 2.62  &  1.69 $\pm$ 1.42  &  107.54 $\pm$ 53.09  \\
        & HKU-IS\cite{li2015visual}  &  4447  &  292.42 $\pm$ 51.13  &  386.64 $\pm$ 37.42  &  488.00 $\pm$ 29.44  &  4.41 $\pm$ 4.28  &  2.21 $\pm$ 2.07  &  114.05 $\pm$ 55.06  \\
        & MSRA-B\cite{DBLP:journals/pami/LiuYSWZTS11}  &  5000  &  321.94 $\pm$ 56.33  &  370.86 $\pm$ 50.84  &  496.42 $\pm$ 22.53  &  2.89 $\pm$ 3.67  &  1.77 $\pm$ 2.25  &  102.04 $\pm$ 56.50  \\
        & DUT-OMRON\cite{yang2013saliency}  &  5168  &  320.93 $\pm$ 54.35  &  376.78 $\pm$ 46.02  &  499.50 $\pm$ 22.97  &  4.08 $\pm$ 6.20  &  2.27 $\pm$ 3.54  &  71.09 $\pm$ 59.60  \\
        & MSRA10K\cite{DBLP:journals/pami/ChengMHTH15}  &  10000  &  324.51 $\pm$ 56.26  &  370.27 $\pm$ 50.25  &  497.57 $\pm$ 22.79  &  2.54 $\pm$ 2.62  &  4.07 $\pm$ 17.94  &  101.95 $\pm$ 63.24  \\ 
        & DUTS\cite{wang2017learning}  &  15572  &  322.1 $\pm$ 53.69  &  375.48 $\pm$ 47.03  &  499.35 $\pm$ 21.95  &  3.37 $\pm$ 4.28  &  2.62 $\pm$ 4.73  &  84.78 $\pm$ 57.74 \\
        & SOC\cite{fan2018SOC}  &  3000  &  480.00 ± 0.00  &  640.00 ± 0.00  &  800.00 ± 0.00  &  4.44 ± 3.57  &  13.69 ± 30.41  &  151.72 ± 154.83  \\
        \hline 
         \multirow{2}{*}{HRS} & HR-SOD\cite{zeng2019towards}  &  2010  &  \ggg 2713.12 \kkk$\pm$ \ggg1041.7\kkk  &  \ggg 3411.81 \kkk$\pm$ \rrr1407.56\kkk  &  \ggg 4405.40 \kkk $\pm$ \rrr1631.03\kkk  &  5.85 $\pm$ 12.60  &  6.33 $\pm$ 16.65  &  319.32 $\pm$ 264.20  \\
        & HR-DAVIS-S\cite{perazzi2016benchmark}  &  92  &  1299.13 $\pm$ 440.77  &  2309.57 $\pm$ 783.59  &  2649.87 $\pm$ 899.05  &  7.84 $\pm$ 5.69  &  15.60 $\pm$ 29.51  &  389.58 $\pm$ 309.29  \\
        \hline 
        \multirow{4}{*}{COD} & CAMO\cite{le2019anabranch}  &  250  &  564.22 $\pm$ 402.12  &  693.89 $\pm$ 578.53  &  905.51 $\pm$ 690.12  &  3.97 $\pm$ 4.47  &  1.48 $\pm$ 1.18  &  65.21 $\pm$ 40.99  \\ 
        & CHAMELEON\cite{chameleon}  &  76  &  741.80 $\pm$ 452.25  &  981.08 $\pm$ 464.88  &  1239.98 $\pm$ 629.19  &  15.25 $\pm$ 51.43  &  10.28 $\pm$ 48.03  &  222.45 $\pm$ 332.22  \\
        & NC4K\cite{fan2020camouflaged}  &  4121  &  529.61 $\pm$ 158.16  &  709.19 $\pm$ 198.90  &  893.23 $\pm$ 223.94  &  7.28 $\pm$ 11.28  &  4.32 $\pm$ 9.44  &  125.43 $\pm$ 123.76  \\
        & COD10K\cite{fan2020camouflaged}  &  5066  &  737.37 $\pm$ 185.65  &  963.85 $\pm$ 222.73  &  1224.53 $\pm$ 239.40  &  \bbb15.28\kkk $\pm$ \bbb71.84\kkk  &  17.18 $\pm$ \bbb183.87\kkk  &  214.12 $\pm$ \bbb857.83\kkk  \\
        \hline 
        \multirow{2}{*}{SMS} & R-PASCAL\cite{cheng2020cascadepsp}  &  501  &  384.34 $\pm$ 64.69  &  469.66 $\pm$ 60.04  &  612.19 $\pm$ 36.32  &  4.44 $\pm$ 6.91  &  7.30 $\pm$ 8.73  &  139.31 $\pm$ 104.60  \\
        & BIG\cite{cheng2020cascadepsp}  &  150  &  \rrr2801.11 \kkk $\pm$ 889.78  &  \rrr 3672.43 \kkk $\pm$ \bbb1128.90\kkk  &  \rrr 4655.81 \kkk $\pm$ \bbb1312.44\kkk  &  11.94 $\pm$ 31.43  &  \bbb31.69\kkk $\pm$ 71.94  &  \ggg655.68 \kkk $\pm$ 710.20  \\
        \hline 
        \multirow{2}{*}{TOS} & COIFT\cite{liew2021deep}  &  280  &  488.27 $\pm$ 92.25  &  600.40 $\pm$ 78.66  &  782.73 $\pm$ 30.45  &  11.88 $\pm$ 12.5  &  4.01 $\pm$ 3.98  &  173.14 $\pm$ 74.54  \\
        & ThinObject5K\cite{liew2021deep}  &  5748  &  1185.59 $\pm$ \bbb909.53\kkk  &  1325.06 $\pm$ 958.43  &  1823.03 $\pm$ 1258.49  &  \ggg26.53 \kkk $\pm$ \ggg119.98\kkk  &  \ggg33.06\kkk $\pm$ \ggg216.07\kkk  &  \bbb519.14 \kkk $\pm$ \ggg1298.54\kkk  \\
        \hline 
        DIS & \textbf{\ourdataset~(Ours)}  &  5470  &  \bbb 2513.37 \kkk $\pm$ \rrr1053.40\kkk  &  \bbb 3111.44 \kkk $\pm$ \ggg1359.51\kkk  &  \bbb 4041.93 \kkk $\pm$ \ggg1618.26\kkk  &  \rrr 107.60 $\pm$ 320.69 \kkk  &  \rrr 106.84 $\pm$ 436.88 \kkk &  \rrr 1427.82 $\pm$ 3326.72 \kkk \\ 
        \bottomrule  
    \end{tabular}
    \vspace{-10pt}
    }
\end{table*}

\noindent\textbf{Data Annotation.} Each image of DIS5K is manually labeled with pixel-wise accuracy using {GIMP}\footnote{\url{https://www.gimp.org/}}. 
The average per-image labeling time is $\sim$30 minutes and some images cost up to 10 hours. It is worth mentioning that some of our labeled ground truth (GT) masks are visually close to the image matting GT. 
The labeled targets, including transparent and translucent, are binary masks with one pixel's highest accuracy. 
Here, the DIS task is category-agnostic while our DIS5K is collected based on 
pre-designed keywords/categories, which seems contradictory. The reasons are threefold. 
(1) The keywords greatly facilitate the retrieval and organization of the large-scale dataset.
(2) To achieve the goal of category-agnostic segmentation, diversified samples are needed. Collecting samples based on their categories is a reasonable way to guarantee the diversities' lower bound of the dataset. The diversities' upper bound of our DIS5K is determined by the diversified characteristics (\eg, textures, structures, shapes, contrasts, complexities, \etc) of a large number of samples, guaranteeing the robustness and generalization of the category-agnostic segmentation. 
(3) There are no perfect datasets, so re-organizing or further extension of the existing datasets is usually necessary for different real-world applications. The category information will significantly facilitate tracing the collected and to-be-collected samples. Therefore, the category-based data collection is not contradictory but internally consistent with the goal of DIS task. 

\subsection{Data Analysis}\label{sec:DA}
For deeper insights into DIS dataset, we compare our \ourdataset~against 
19 other related datasets including: 
(1) nine salient object detection (SOD) datasets: SOD~\cite{movahedi2010design}, PASCAL-S \cite{li2014secrets}, ECSSD~\cite{yan2013hierarchical},
HKU-IS \cite{li2015visual}, MSRA-B \cite{DBLP:journals/pami/LiuYSWZTS11}, 
DUT-OMRON \cite{yang2013saliency}, MSRA10K \cite{DBLP:journals/pami/ChengMHTH15}, DUTS~\cite{wang2017learning}, and SOC~\cite{fan2018SOC}; 
(2) two high-resolution salient object detection (HR-SOD) datasets: HR-SOD \cite{zeng2019towards} and HR-DAVIS-S \cite{perazzi2016benchmark,zeng2019towards}; 
(3) four camouflaged object detection (COD) datasets: CAMO \cite{le2019anabranch}, CHAMELEON \cite{chameleon}, COD10K~\cite{fan2020camouflaged}, and NC4K~\cite{lv2021simultaneously}; 
(4) two semantic segmentation (SMS)\footnote{
It is worth noting that only R-PASCAL and the BIG datasets are included here because they target highly accurate segmentation, and most of their images contain one or two objects, which is comparable to the listed tasks and datasets.} datasets: R-PASCAL \cite{everingham2010pascal,cheng2020cascadepsp} and BIG \cite{cheng2020cascadepsp}; 
(5) two thin object segmentation (TOS) datasets: COIFT \cite{liew2021deep} and ThinObject5K~\cite{liew2021deep}. 
The comparisons are conducted mainly from the following three perspectives: \textit{image number}, \textit{image dimension}, and \textit{object complexity} as illustrated in \tabref{tab:my_label}

\noindent\textbf{Image Dimension} is crucial to segmentation tasks. Because it has significant impacts on accuracy, efficiency, and computational costs. The mean ($H$, $W$, $D$) and their standard deviations ($\sigma_H$, $\sigma_W$, $\sigma_D$) of the image height, width and diagonal length are provided in \tabref{tab:my_label}. The BIG dataset has the largest average image dimensions, but it is a small-scale dataset that contains only 150 images.
Although HR-SOD has slightly greater dimensions than ours, the dataset scale and complexity are less comparable. Compared with the SOD datasets, the average image dimensions of our DIS5K are almost eight times larger than theirs. 
The COD datasets have larger dimensions than SOD datasets, 
but they are still much smaller than ours. 
Besides, most of the targets in COD datasets are animals. 
Thus, it is difficult to generalize them to diversified tasks.

\noindent\textbf{Object Complexity} is described by three metrics including the \textit{isoperimetric inequality quotient} ($IPQ$) \cite{osserman1978isoperimetric,watson2012perimetric,yang2020meticulous}, the \textit{number of object contours} ($C_{num}$) and the \textit{number of dominant points} $P_{num}$. The $IPQ$ mainly describes the overall structure complexity as $IPQ = \frac{L^2}{4\pi A}$, where $L$ and $A$ denote the object perimeter and the region area, respectively. 
It is designed to differentiate objects with elongated components and thin concave structures from close-to-convex objects. The $C_{num}$ is used to represent the topological complexity in contour level for observing the objects consisting of many (small) contours which usually have minor influences on the $IPQ$. To describe the object complexity at a finer level, we employ $P_{num}$ to count the number of the dominant points \cite{DBLP:journals/cvgip/Ramer72} along the object boundaries. 
Therefore, the complexities of the small jagged segments along the boundaries, which usually cannot be accurately measured by $IPQ$ and $C_{num}$, can be well-evaluated with $P_{num}$. Essentially, $P_{num}$ is the total number of the polygon corners needed for approximating the segmentation masks, which also directly reflects the human labeling costs. Thus, it is then adapted to our Human Correction Efforts (HCE) metric (\secref{sec:HCE}) for evaluating the prediction quality.

\begin{figure*}[t!]
    \centering
    \includegraphics[width=\linewidth]{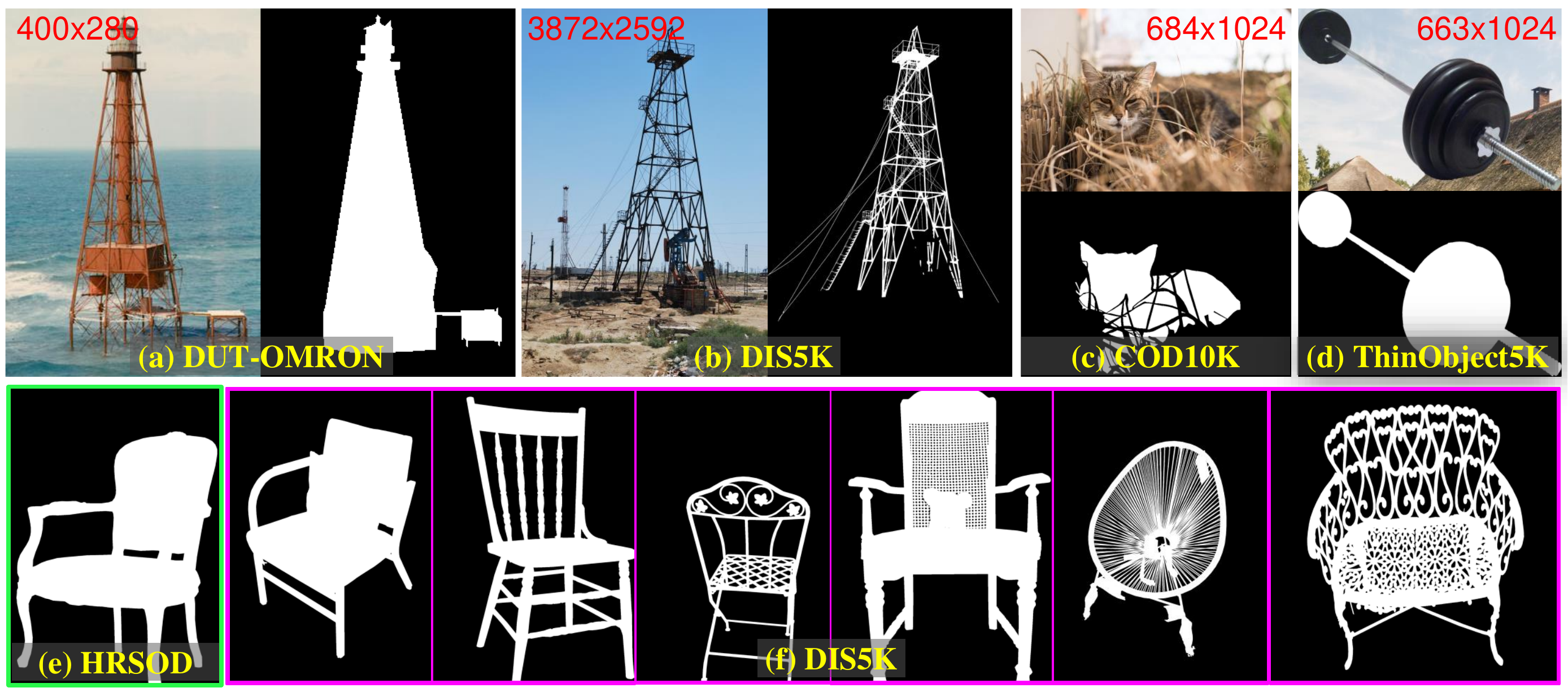}
    \vspace{-20pt}
    \caption{\small Qualitative comparisons of different datasets. (a) and (b) indicate that our DIS5K provides more accurate labels. (c) shows one sample from COD10K~\cite{fan2020camouflaged}, of which the structural complexity is caused by occlusion. (d) illustrates the synthetic ThinObject5K~\cite{liew2021deep} dataset. (e) and (f) demonstrate that DIS5K has a larger diversity of intra-categorical structure complexities.}
    \label{fig:datasets-qual}
    \vspace{-10pt}
\end{figure*}

\begin{figure}[t!]
    \centering
    \includegraphics[width=\linewidth]{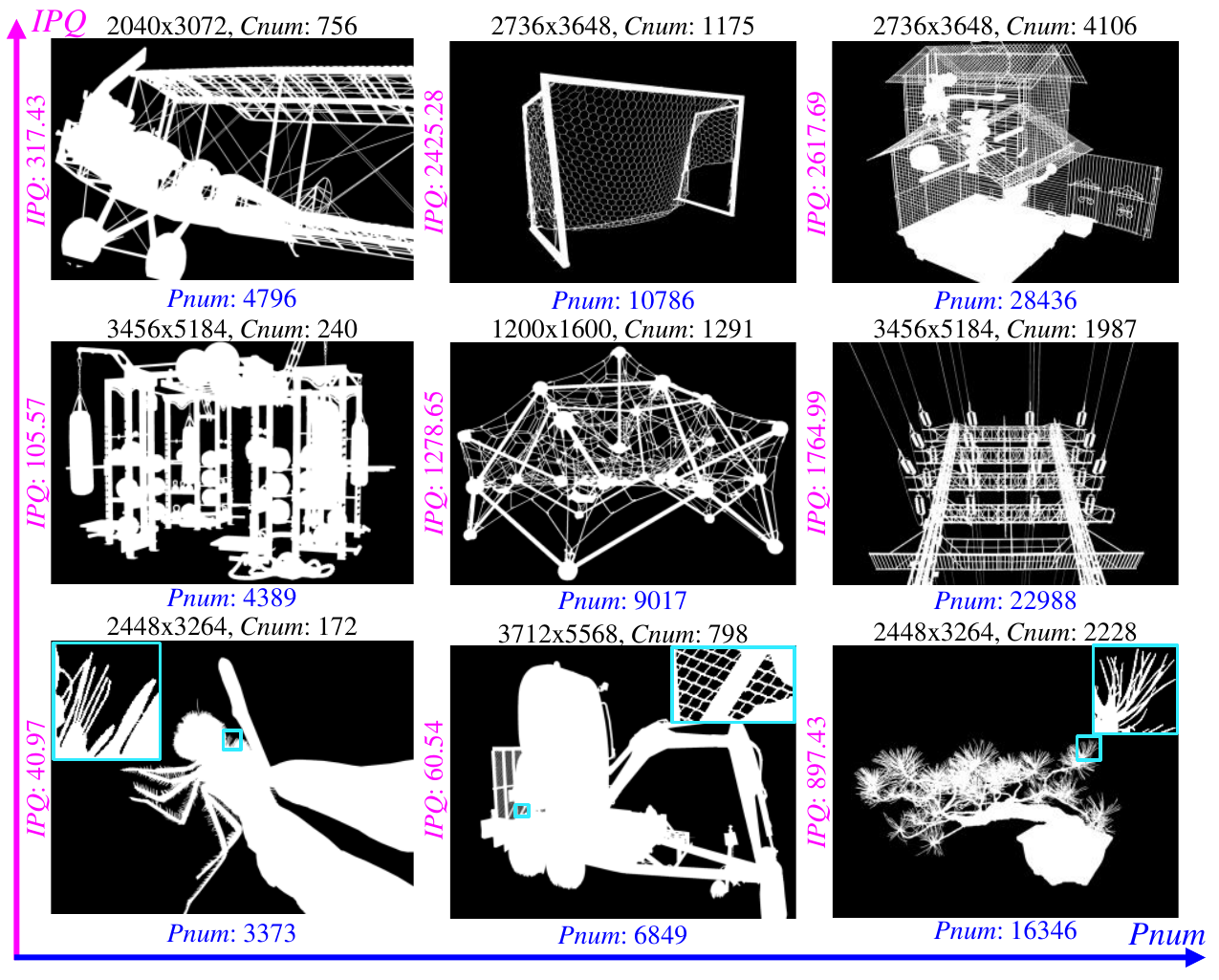}
    \vspace{-10pt}
    \caption{\small GT masks of our DIS5K with diversified inter-categorical complexities. The complexity relationships are only valid within each row or column.}
    \label{fig:datasets-cmplx}
    \vspace{-10pt}
\end{figure}

\noindent\textbf{Discussion.} The metrics above are all computed on the labeled GT masks and illustrated in \tabref{tab:my_label} and \figref{fig:DIS-cats-grps} (Left). It shows that DIS5K is around 20 (up to 50) times more complicated than the SOD datasets in terms of average structure complexity $IPQ$. Although other datasets such as CHAMELEON, COD10K, BIG, COIFT, and ThinObject5K have higher average $IPQ$ against the SOD datasets, their complexities are still much less than ours. The HR-SOD and HR-DAVIS-S datasets contain large-size images with accurately labeled boundaries. However, there are no significant differences between their $IPQ$ and that of SOD datasets. Because $IPQ$ is insensitive to the complexities of fine details as mentioned above. The average contour-level complexities $C_{num}$ of different datasets are almost consistent with their $IPQ$. The average $C_{num}$ and its standard deviation of DIS5K are over 100 and 400, which are much higher than other datasets. This indicates the objects in DIS5K contain more detailed structures that are comprised of multiple contours. The average $P_{num}$ of DIS5K is over 1400, which is almost five and three times greater than those of HR-SOD and the synthetic ThinObject5K, respectively. There is an interesting observation that the $P_{num}$ of HR-SOD, HR-DAVIS-S, BIG, and ThinObject5K are not proportional to their $IPQ$ and $C_{num}$, but it shows positive correlations with their image dimensions. One of the reasons is that most of the objects in these datasets are close to convex and comprised of single or a few contours, which leads to low $IPQ$ and $C_{num}$.
Nevertheless, their boundaries (\eg, small jagged segments) are accurately labeled in high-resolution images that significantly increase the $P_{num}$. 
On the other hand, larger sizes of GT masks often directly lead to greater $P_{num}$ because the dominant points are searched by \cite{DBLP:journals/cvgip/Ramer72}, which filters out redundant boundary points based on their deviation distances ($epsilon$) against the straight lines constructed by their neighboring dominant points. For example, given two objects with the same shape comprised of smooth boundaries but different sizes, more dominant points are generated from the larger one with the same threshold of $epsilon$. That means $P_{num}$ is determined by both the boundary complexity and the GT mask dimension. Therefore, these three complexity measurements are complementary to provide a comprehensive analysis of the object complexities.The large standard deviations in \tabref{tab:my_label} demonstrate the great diversities of DIS5K from different perspectives. 

\begin{figure*}[t!]
    \centering
    \includegraphics[width=.8\textwidth]{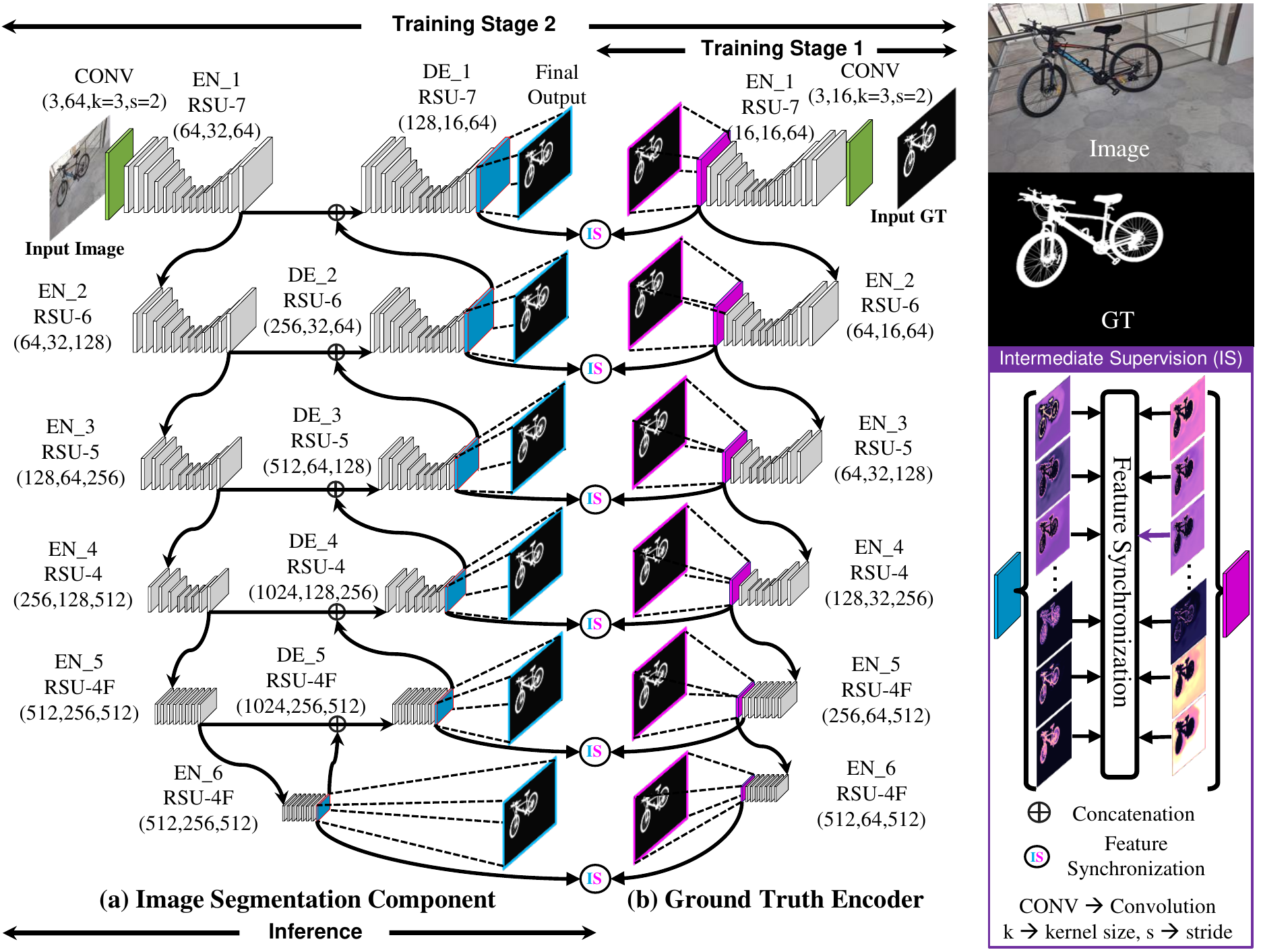}
    \vspace{-10pt}
    \caption{\small 
    Proposed \ourmodel~baseline: (a) shows the image segmentation component, (b) illustrates the ground truth encoder built upon the intermediate supervision (IS) component.}
    \label{fig:model}
    \vspace{-10pt}
\end{figure*}

\figref{fig:datasets-qual}-a shows an observation tower from DUT-OMRON. Similar object (b) has also been included in our DIS5K, which has higher labeling accuracy and structural complexity.  \figref{fig:datasets-qual}-c shows a sample from COD10K where the relatively higher structure complexity than that of SOD datasets is partially caused by the labeled occlusions, which are not the structural complexity of the target itself. A sample, where a set of the barbell is floating in the sky, from the synthesized ThinObject5K dataset is shown in \figref{fig:datasets-qual}-d. Synthesizing images is a common way for generating training sets in image matting \cite{xu2017deep,yu2020high}, where training samples are difficult to be labeled. However, the synthesized images usually show different characteristics from the real ones, which leads to biases in both training and evaluation. \figref{fig:datasets-qual}-e and \figref{fig:datasets-qual}-f demonstrate the larger diversity of intra-categorical structure complexities of our DIS5K. In \figref{fig:datasets-cmplx}, we provide the sample masks with their complexity scores in DIS5K. The bottom-left samples with large regional components have relatively low $IPQ$, and the top-right samples with more thin and complicated fine structures have much higher $IPQ$ and $P_{num}$. 

\subsection{Dataset Splitting}
We split 5,470 images in DIS5K into three subsets: DIS-TR (3,000), DIS-VD (470), and DIS-TE (2,000) for training, validation, and testing. The categories in DIS-TR and those in DIS-VD and DIS-TE are mainly consistent. Since our dataset's object shapes and structure complexities are diversified, the 2000 images of DIS-TE are further split into four subsets with ascending shape complexities for a more comprehensive evaluation. Specifically, we first rank the 2,000 testing images in ascending order according to the multiplication ($IPQ\times P_{num}$) of their structure complexities $IPQ$ and boundary complexities $P_{num}$. Then, DIS-TE is split into four subsets (DIS-TE1$\sim$DIS-TE4) with 500 images in each subset to represent four testing difficulty levels.


\begin{figure*}[t!]
    \centering
    \includegraphics[width=.98\textwidth]{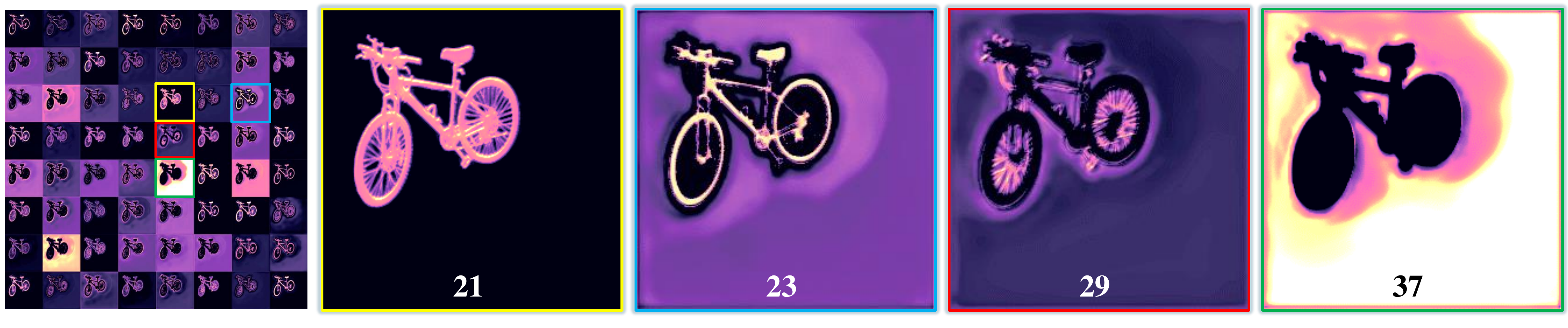}
    \vspace{-10pt}
    \caption{\small 
    Feature maps produced by the last layer of the EN\_2 stage of our GT encoder. ``21'', ``23'', ``29'' and ``37'' are the indices (start with 1) of the corresponding channels in the feature map.}
    \label{fig:FS-features}
    \vspace{-10pt}
\end{figure*}

\section{Proposed \ourmodel~Baseline}
\subsection{Overview}
As shown in \figref{fig:model}, our \ourmodel~consists of a ground truth (GT) encoder, a image segmentation component, and a newly proposed intermediate supervision strategy.
The \textbf{GT encoder} (27.7 MB) is designed to encode the GT masks into high-dimensional spaces and then used to enforce intermediate supervision on the segmentation component. While, 
the \textbf{image segmentation component} (176.6 MB) is expected to have the capability of capturing fine structures and handle large size \eg, $1024\times1024$, inputs with affordable memory and time costs. In the following experiment, we choose U$^2$-Net~\cite{qin2020u2} as the image segmentation component because of its strong capability in capturing fine structures. Note that other segmentation models, such as transformer backbone, are also compatible with our strategy.

\noindent\textbf{Technique Details.}
U$^2$-Net was originally designed for small size ($320\times 320$) SOD image. Because of its GPU memory costs, it cannot be used directly for handling large size (\eg, $1024\times 1024$) inputs. We adapt the architecture of U$^2$-Net by adding an 
input convolution layer before its first encoder stage. 
The input convolution layer is set as a plain convolution layer with a kernel size of $3\times3$ and stride of $2$. Given an input image with a shape of $I^{1024\times1024\times3}$, the input convolution layer first transforms it to a feature map $f^{512\times512\times64}$ and this feature map is then directly
fed to the original U$^2$-Net, where the input channel is changed to 64 correspondingly. Compared with directly feeding $I^{1024\times1024\times3}$ to U$^2$-Net, the input convolution layer helps the whole network reduce three quarters of the overall GPU memory overhead while maintaining spatial information in feature channels. 

\subsection{Intermediate Supervision}
DIS can be seen as a mapping in segmentation models from image domain $\mathcal{I} \in \mathbb{R}^{H \times W \times 3}$ to segmentation GT domain $\mathcal{G} \in \mathbb{R}^{H \times W \times 1}$: $\mathcal{G} = F(\theta, \mathcal{I}),$ where $F$ indicates the model that uses learnable weights $\theta$ to map inputs from image to mask domain. 
Most of the models are easy to over-fit on the training set. Thus, the deep supervision has been proposed to supervise the intermediate outputs of a given deep network \cite{lee2015deeply}. 
In~\cite{xie2015holistically,qin2020u2}, the dense supervisions are usually applied to the side outputs, which are single-channel probability maps produced by convolving the last feature maps of particular deep layers. 
However, transforming high-dimensional features to single-channel probability maps is essentially a dimension reduction operation, inevitably losing critical cues. 

To avoid this issue, we propose a novel intermediate supervision 
training strategy. Given an input image $I^{H \times W \times 3}$ and its corresponding segmentation mask $G^{W \times H \times 1}$, we first train a self-supervised GT encoder to extract the high-dimensional features using a lightweight deep model $F_{gt}$, \figref{fig:model}-b, as: 
$\underset{\theta_{gt}}{\mathrm{argmin}}\sum_{d=1}^{D}BCE(F_{gt}(\theta_{gt},G)_d,G)$, where $\theta_{gt}$ indicates the model weights, $BCE$ is the binary cross entropy loss and $D$ denotes the number of the intermediate feature maps. 

After obtaining the GT encoder $F_{gt}$, its weights $\theta_{gt}$ are frozen for generating the ``ground truth'' high-dimensional intermediate deep features by:
$f_D^G = F_{gt}^-(\theta_{gt}, G), D = \{1, 2, 3, 4, 5, 6\}$, 
where $F_{gt}^-$ represents the $F_{gt}$ without the last convolution layers for generating the probability maps. $F_{gt}^-$ is to supervise those corresponding features $f_D^I$ from the segmentation model $F_{sg}$. In the image segmentation component $F_{sg}$ (\figref{fig:model}-a), the image $I$ is transformed to a set of high-dimensional intermediate feature maps $f_D^I$ before producing the probability maps. Each feature map $f_d^I$ has the same dimension with its corresponding GT intermediate feature map $f_d^G$: 
$f_D^I = F_{sg}^-(\theta_{sg}, I), D = \{1, 2, 3, 4, 5, 6\}$, 
where $\theta_{sg}$ denotes the weights of the segmentation model. Then, the intermediate supervision (IS) via \textit{feature synchronization} on the deep intermediate features can be conducted by the following high-dimensional feature consistency loss: 
$L_{\text{fs}} = \sum_{d=1}^{D}\lambda^{fs}_d\left\|f_{d}^{I}-f_{d}^{G}\right\|^{2}$, 
where $\lambda^{fs}_d$ denotes the weight of each FS loss. 
The training process of the segmentation model $F_{sg}$ can be formulated as the following optimization problem:
$\underset{\theta_{sg}}{\mathrm{argmin}}(L_{\text{fs}} + L_{\text{sg}})$, 
where $L_{\text{sg}}$ indicates the $BCE$ loss of the side outputs of $F_{sg}$: 
$L_{\text{sg}} = \sum_{d=1}^{D}\lambda^{sg}_d{BCE(F_{sg}(\theta_{sg}, I), G)}$,
where $\lambda^{sg}_d$ represents the hyperparameter to weight each side output loss. 

\figref{fig:FS-features} illustrates the feature maps from the stage 2 in \figref{fig:model}, EN\_2, of the GT encoder. We can see the diversified characteristics of the input mask are encoded into different channels. For example, the 21$^{st}$ channel encodes both the fine and large structures close to the original mask. While the 23$^{rd}$, 29$^{th}$, and 37$^{th}$ channels encode the middle size structures (frame, seat, wheels), delicate structures (brake cables and spokes), large size region (the overall shape of the bicycle), respectively. These diversified features of the GT can provide stronger regularizations and more comprehensive supervisions for reducing the risks of over-fitting.

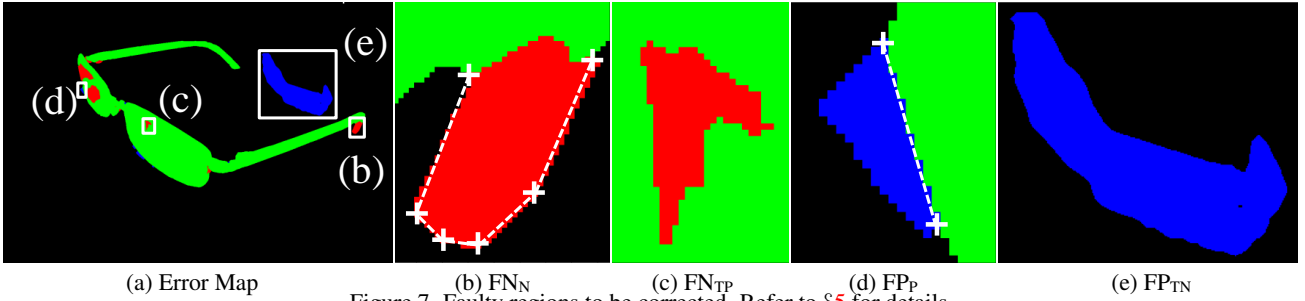
\begin{figure*}[t!]
    \centering
    \begin{overpic}[width=\textwidth]{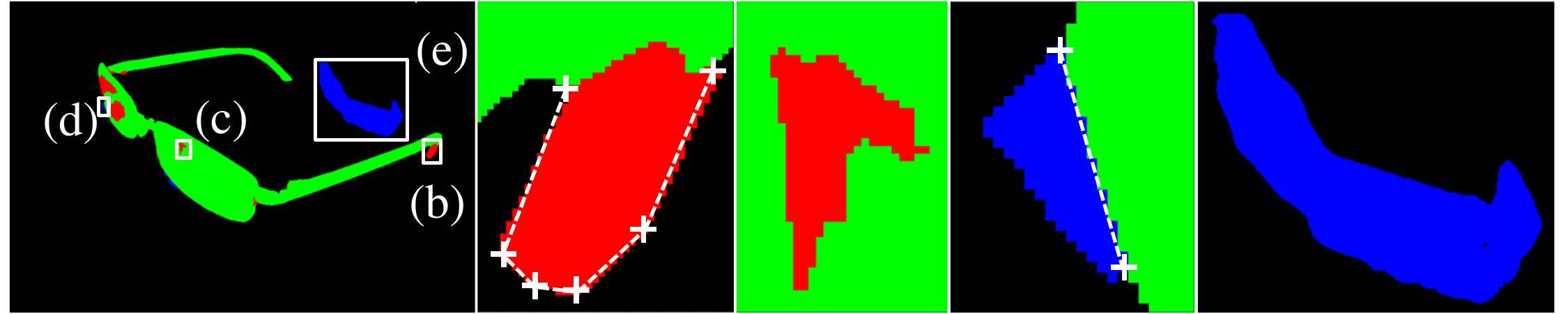}
    \put(10,-2.0) {\small (a) Error Map}
    \put(35,-2.0) {\small (b) FN$_\text{N}$}
    \put(50,-2.0) {\small (c) FN$_\text{TP}$}
    \put(65,-2.0) {\small (d) FP$_\text{P}$}
    \put(85,-2.0) {\small (e) FP$_\text{TN}$}
    \end{overpic}
    \caption{\small
    Faulty regions to be corrected. Refer to \secref{sec:HCE} for details.
    }\label{fig:hce}
    \vspace{-12pt}
\end{figure*}

\section{Proposed HCE Metric}\label{sec:HCE}
Given a predicted segmentation probability map $P \in \mathbb{R}^{W \times H \times 1}$ and its corresponding GT mask $G \in \mathbb{R}^{W \times H \times 1}$, the existing metrics, \eg, IoU, boundary IoU \cite{DBLP:conf/cvpr/ChengGDBK21}, F-measure \cite{achanta2009frequency}, boundary F-measure \cite{ehrig2005relaxed,qin2019basnet}, and 
MAE \cite{perazzi2012saliency}, usually evaluate the quality of the prediction $P$ by calculating the scores based on the mathematical or cognitive consistency (or inconsistency) between $P$ and $G$.
In other words, these metrics describe how significant the ``gap'' is between $P$ and $M$ from different perspectives.
However, measuring the magnitude of the ``gap'' is insufficient when applying the models in many real-world applications, where evaluating the costs of filling the ``gap'' is more important.

Therefore, we propose a novel evaluation metric, Human Correction Efforts (HCE), which approximates the human efforts required in correcting faulty predictions to satisfy specific accuracy requirements in real-world applications. According to our labeling experiences, there are mainly two frequently used operations: (1) points selection along target boundaries to formulate polygons and (2) region selection based on similar pixel intensities inside the region. Both operations correspond to one mouse click by the human operator. Therefore, the HCE here is quantified by the approximated number of mouse clicking numbers. 
Particularly, to correct a faulty predicted mask, the operators need to manually sample dominant points along the erroneously predicted targets' boundaries or regions for correcting both False Positive (FP) and False Negative (FN) regions. 
As shown in \figref{fig:hce}, the FNs and FPs can be categorized into two classes, respectively, according to their adjacent regions: FN$_{\text{N}}$ (N=TN+FP), FN$_{\text{TP}}$, FP$_{\text{P}}$ (P=TP+FN) and FP$_{\text{TN}}$. 
To correct the FN$_{\text{N}}$ regions, its boundaries adjacent to the TN need to be manually labeled with dominant points (\figref{fig:hce}-b). Similarly, to correct the FP$_{\text{P}}$ regions, we only need to label its boundaries adjacent to the TP regions (\figref{fig:hce}-d). The FN$_{\text{TP}}$ regions (\figref{fig:hce}-c) enclosed by TP and the FP$_{\text{TN}}$ regions (\figref{fig:hce}-e) enclosed by TN can be easily corrected by one-click region selection. Therefore, the HCE for correcting the faulty regions in \figref{fig:hce} (b-e) is 10 (six and two clicks needed in (b) and (d), one click needed in (c) and one click needed in (e)). The dominant point selection operations and the region selection operations are approximated by DP algorithm \cite{DBLP:journals/cvgip/Ramer72} based on the contours obtained by OpenCV findContours \cite{DBLP:journals/cvgip/SuzukiA85} function and the connected regions labeling algorithm \cite{DBLP:journals/tcs/FiorioG96,DBLP:conf/miip/WuOS05}, respectively, in the evaluation stage. 

\begin{algorithm}[t!]
\scriptsize
\KwIn{$P$,~$G$,~$\gamma=5$,~$epsilon=2.0$}
\KwOut{$HCE_{\gamma}$}
$G_{ske}$ = skeletonize ($G$)\;
$P_{or}G$,~$TP$ = or ($P$, $G$),~and ($P$,$G$)\;
$FN$, $FP$=  ($G$ - $TP$), ($P$ - $TP$)\;
\For{($i=0;i\le\gamma;i++$)}{
    $P_{or}G$ = erode ($P_{or}G$,~$disk~(1)$)\;
}
$FN'$, $FP'$ = and ($FN$,$P_{or}G$), and ($FP$,$P_{or}G$)\;
\For{($i=0;i\le\gamma;i++$)}{
    $FN'$ = dilate ($FN'$,~$disk~(1)$)\;
    $FN'$ = and ($FN'$, not $P$)\;
    $FP'$ = dilate ($FP'$,~$disk~(1)$)\;
    $FP'$ = and ($FP'$, not $G$)\;
}
$FN'$, $FP'$ = and ($FN$,~$FN'$), and ($FP$,~$FP'$)\;
$FN'$ = or ($FN'$, xor ($G_{ske}$, and ($TP$,~$G_{ske}$)))\;
$HCE_{\gamma}$ = compute\_HCE ($FN'$, $FP'$,~$TP$, epsilon)
\caption{\small Relax HCE.}
\label{alg:relax_hce}

\end{algorithm}

\noindent\textbf{Relax HCE.} 
Practically, some applications may be tolerant to certain minor prediction errors. Therefore, we extend the computation of HCE by taking the error tolerance $\gamma$ into consideration ($HCE_{\gamma}$). The key idea is to relax the FP and FN regions by excluding the small FP and FN components using erosion \cite{haralick1987image} and dilation \cite{haralick1987image} operations. Given a segmentation map $P$, its corresponding GT mask $G$, the error tolerance (\eg, $\gamma=5$, which denotes the size of the to-be-ignored small faulty regions), the $epsilon$ of DP algorithm, the computation of the $HCE_{\gamma}$ can be summarized in Alg.~\ref{alg:relax_hce}. Note that the erosion operation (Line5 of Alg.~\ref{alg:relax_hce}) can remove all thin and fine components of $P_{or}G$. 
However, some of the thin components (\eg, thin cables, nets) are critical in representing the targets, and they need to be retained regardless of their sizes.  
To address this, the skeleton of the GT mask is extracted by \cite{DBLP:journals/cacm/ZhangS84} and combined with the relaxed $FN'$ mask for retaining these structures. 




\begin{table*}[t!]
\centering
	\caption{\small Quantitative evaluation on \ourdataset~validation and test sets. R = ResNet~\cite{he2016deep}. R2 = Res2Net~\cite{gao2019res2net}. S-813 = STDC813~\cite{Fan_2021_CVPR}, E-B1 = EffinetB1~\cite{tan2019efficientnet}.
	}\label{tab:compSOA}
    \resizebox{\textwidth}{!}{
	\begin{tabular}{c|r|cccccc|cc|ccc|ccccc|c}
		\hline
        Dataset	&	Metric	&	\tabincell{c}{UNet\\\cite{ronneberger2015u}}	&	\tabincell{c}{BASNet\\\cite{qin2019basnet}}	&	\tabincell{c}{GateNet\\\cite{zhao2020suppress}}	&	\tabincell{c}{F$^3$Net\\\cite{wei2020f3net}}	&	\tabincell{c}{GCPANet\\\cite{chen2020global}}	&	\tabincell{c}{U$^2$Net\\\cite{qin2020u2}}	&	\tabincell{c}{SINetV2\\\cite{fan2021concealed}}	&	\tabincell{c}{PFNet\\\cite{Mei_2021_CVPR}}	&	\tabincell{c}{PSPNet\\\cite{zhao2017pyramid}}	&	\tabincell{c}{DLV3+\\\cite{chen2018encoder}}	&
        \tabincell{c}{HRNet\\\cite{WangSCJDZLMTWLX19}} &
        \tabincell{c}{BSV1\\\cite{yu2018bisenet}}	&	\tabincell{c}{ICNet\\\cite{zhao2018icnet}}	&
        \tabincell{c}{MBV3\\\cite{DBLP:conf/iccv/HowardPALSCWCTC19}} & \tabincell{c}{STDC\\\cite{Fan_2021_CVPR}}	&	\tabincell{c}{HySM\\\cite{nirkin2020hyperseg}}	&	\textbf{\ourmodel}	\\
        \hline 
        \multirow{4}{*}{\begin{sideways}\tabincell{c}{\textbf{Attr.}}\end{sideways}}	&	Backbone	&	-	&	R-34	&	R-50	&	R-50	&	R-50	&	-	&	R2-50	&	R-50	&	R-50	&	R-50	&	-	&	R-18	&	R-18	& MBV3	& S-813	&	E-B1	&	-	\\
        	&	Size (MB)	&	121.4	&	348.6	&	515.0	&	102.6	&	268.7	&	176.3	&	108.5	&	186.6	&	196.1	&	161.8	&	264.4	&	47.6	&	46.5 & 21.5	&	48.4	&	49.6	&	176.6	\\
        &	Time (ms) &	3.87	&	10.71	&	12.69	&	14.23	&	11.04	&	19.73	&	18.69	&	17.16	&	8.08	&	8.68	&	40.5	&	6.07	&	4.93 & 8.86	&	6.17	&	24.06	&	19.49	\\
        	&	Input Size	&	$512^2$	&	$320^2$	&	$384^2$	&	$352^2$	&	$320^2$	&	$320^2$	&	$352^2$	&	$416^2$	&	$512^2$	&	$513^2$	&	$1024^2$	&	\tiny{1024x2048}	&	\tiny{1024x2048} & $1024^2$	&	\tiny{512x1024}	&	\tiny{512x1024}	&	$1024^2$	\\
        \hline 
        \multirow{6}{*}{\begin{sideways}\textbf{DIS-VD}\end{sideways}}	&	$maxF_\beta\uparrow$	&	0.692	&	0.731	&	0.678	&	0.685	&	0.648	&	0.748	&	0.665	&	0.691	&	0.691	&	0.660	&	0.726	&	0.662	&	0.697	&	0.714	&	0.696	&	0.734	&	\textbf{0.791}	\\
	&	$F^w_\beta\uparrow$	&	0.586	&	0.641	&	0.574	&	0.595	&	0.542	&	0.656	&	0.584	&	0.604	&	0.603	&	0.568	&	0.641	&	0.548	&	0.609	&	0.642	&	0.613	&	0.640	&	\textbf{0.717}	\\
	&	$~M~\downarrow$	&	0.113	&	0.094	&	0.110	&	0.107	&	0.118	&	0.090	&	0.110	&	0.106	&	0.102	&	0.114	&	0.095	&	0.116	&	0.102	&	0.092	&	0.103	&	0.096	&	\textbf{0.074}	\\
	&	$S_{\alpha}\uparrow$	&	0.745	&	0.768	&	0.723	&	0.733	&	0.718	&	0.781	&	0.727	&	0.740	&	0.744	&	0.716	&	0.767	&	0.728	&	0.747	&	0.758	&	0.740	&	0.773	&	\textbf{0.813}	\\
	&	$E_{\phi}^{m}\uparrow$	&	0.785	&	0.816	&	0.783	&	0.800	&	0.765	&	0.823	&	0.798	&	0.811	&	0.802	&	0.796	&	0.824	&	0.767	&	0.811	&	0.841	&	0.817	&	0.814	&	\textbf{0.856}	\\
	&	$HCE_\gamma\downarrow$	&	1337	&	1402	&	1493	&	1567	&	1555	&	1413	&	1568	&	1606	&	1588	&	1520	&	1560	&	1660	&	1503	&	1625	&	1598	&	1324	&	\textbf{1116}	\\
	\hline 
\multirow{6}{*}{\begin{sideways}\textbf{DIS-TE1}\end{sideways}}	&	$maxF_\beta\uparrow$	&	0.625	&	0.688	&	0.620	&	0.640	&	0.598	&	0.694	&	0.644	&	0.646	&	0.645	&	0.601	&	0.668	&	0.595	&	0.631	&	0.669	&	0.648	&	0.695	&	\textbf{0.740}	\\
	&	$F^w_\beta\uparrow$	&	0.514	&	0.595	&	0.517	&	0.549	&	0.495	&	0.601	&	0.558	&	0.552	&	0.557	&	0.506	&	0.579	&	0.474	&	0.535	&	0.595	&	0.562	&	0.597	&	\textbf{0.662}	\\
	&	$~M~\downarrow$	&	0.106	&	0.084	&	0.099	&	0.095	&	0.103	&	0.083	&	0.094	&	0.094	&	0.089	&	0.102	&	0.088	&	0.108	&	0.095	&	0.083	&	0.090	&	0.082	&	\textbf{0.074}	\\
	&	$S_{\alpha}\uparrow$	&	0.716	&	0.754	&	0.701	&	0.721	&	0.705	&	0.760	&	0.727	&	0.722	&	0.725	&	0.694	&	0.742	&	0.695	&	0.716	&	0.740	&	0.723	&	0.761	&	\textbf{0.787}	\\
	&	$E_{\phi}^{m}\uparrow$	&	0.750	&	0.801	&	0.766	&	0.783	&	0.750	&	0.801	&	0.791	&	0.786	&	0.791	&	0.772	&	0.797	&	0.741	&	0.784	&	0.818	&	0.798	&	0.803	&	\textbf{0.820}	\\
	&	$HCE_\gamma\downarrow$	&	233	&	220	&	230	&	244	&	271	&	224	&	274	&	253	&	267	&	234	&	262	&	288	&	234	&	274	&	249	&	205	&	\textbf{149}	\\
	\hline 
\multirow{6}{*}{\begin{sideways}\textbf{DIS-TE2}\end{sideways}}	&	$maxF_\beta\uparrow$	&	0.703	&	0.755	&	0.702	&	0.712	&	0.673	&	0.756	&	0.700	&	0.720	&	0.724	&	0.681	&	0.747	&	0.680	&	0.716	&	0.743	&	0.720	&	0.759	&	\textbf{0.799}	\\
	&	$F^w_\beta\uparrow$	&	0.597	&	0.668	&	0.598	&	0.620	&	0.570	&	0.668	&	0.618	&	0.633	&	0.636	&	0.587	&	0.664	&	0.564	&	0.627	&	0.672	&	0.636	&	0.667	&	\textbf{0.728}	\\
	&	$~M~\downarrow$	&	0.107	&	0.084	&	0.102	&	0.097	&	0.109	&	0.085	&	0.099	&	0.096	&	0.092	&	0.105	&	0.087	&	0.111	&	0.095	&	0.083	&	0.092	&	0.085	&	\textbf{0.070}	\\
	&	$S_{\alpha}\uparrow$	&	0.755	&	0.786	&	0.737	&	0.755	&	0.735	&	0.788	&	0.753	&	0.761	&	0.763	&	0.729	&	0.784	&	0.740	&	0.759	&	0.777	&	0.759	&	0.794	&	\textbf{0.823}	\\
	&	$E_{\phi}^{m}\uparrow$	&	0.796	&	0.836	&	0.804	&	0.820	&	0.786	&	0.833	&	0.823	&	0.829	&	0.828	&	0.813	&	0.840	&	0.781	&	0.826	&	0.856	&	0.834	&	0.832	&	\textbf{0.858}	\\
    &	$HCE_\gamma\downarrow$	&	474	&	480	&	501	&	542	&	574	&	490	&	593	&	567	&	586	&	516	&	555	&	621	&	512	&	600	&	556	&	451	&	\textbf{340}	\\
    \hline 
\multirow{6}{*}{\begin{sideways}\textbf{DIS-TE3}\end{sideways}}	&	$maxF_\beta\uparrow$	&	0.748	&	0.785	&	0.726	&	0.743	&	0.699	&	0.798	&	0.730	&	0.751	&	0.747	&	0.717	&	0.784	&	0.710	&	0.752	&	0.772	&	0.745	&	0.792	&	\textbf{0.830}	\\
	&	$F^w_\beta\uparrow$	&	0.644	&	0.696	&	0.620	&	0.656	&	0.590	&	0.707	&	0.641	&	0.664	&	0.657	&	0.623	&	0.700	&	0.595	&	0.664	&	0.702	&	0.662	&	0.701	&	\textbf{0.758}	\\
	&	$~M~\downarrow$	&	0.098	&	0.083	&	0.103	&	0.092	&	0.109	&	0.079	&	0.096	&	0.092	&	0.092	&	0.102	&	0.080	&	0.109	&	0.091	&	0.078	&	0.090	&	0.079	&	\textbf{0.064}	\\
	&	$S_{\alpha}\uparrow$	&	0.780	&	0.798	&	0.747	&	0.773	&	0.748	&	0.809	&	0.766	&	0.777	&	0.774	&	0.749	&	0.805	&	0.757	&	0.780	&	0.794	&	0.771	&	0.811	&	\textbf{0.836}	\\
	&	$E_{\phi}^{m}\uparrow$	&	0.827	&	0.856	&	0.815	&	0.848	&	0.801	&	0.858	&	0.849	&	0.854	&	0.843	&	0.833	&	0.869	&	0.801	&	0.852	&	0.880	&	0.855	&	0.857	&	\textbf{0.883}	\\
	&	$HCE_\gamma\downarrow$	&	883	&	948	&	972	&	1059	&	1058	&	965	&	1096	&	1082	&	1111	&	999	&	1049	&	1146	&	1001	&	1136	&	1081	&	887	&	\textbf{687}	\\
	\hline 
\multirow{6}{*}{\begin{sideways}\textbf{DIS-TE4}\end{sideways}}	&	$maxF_\beta\uparrow$	&	0.759	&	0.780	&	0.729	&	0.721	&	0.670	&	0.795	&	0.699	&	0.731	&	0.725	&	0.715	&	0.772	&	0.710	&	0.749	&	0.736	&	0.731	&	0.782	&	\textbf{0.827}	\\
	&	$F^w_\beta\uparrow$	&	0.659	&	0.693	&	0.625	&	0.633	&	0.559	&	0.705	&	0.616	&	0.647	&	0.630	&	0.621	&	0.687	&	0.598	&	0.663	&	0.664	&	0.652	&	0.693	&	\textbf{0.753}	\\
	&	$~M~\downarrow$	&	0.102	&	0.091	&	0.109	&	0.107	&	0.127	&	0.087	&	0.113	&	0.107	&	0.107	&	0.111	&	0.092	&	0.114	&	0.099	&	0.098	&	0.102	&	0.091	&	\textbf{0.072}	\\
	&	$S_{\alpha}\uparrow$	&	0.784	&	0.794	&	0.743	&	0.752	&	0.723	&	0.807	&	0.744	&	0.763	&	0.758	&	0.744	&	0.792	&	0.755	&	0.776	&	0.770	&	0.762	&	0.802	&	\textbf{0.830}	\\
	&	$E_{\phi}^{m}\uparrow$	&	0.821	&	0.848	&	0.803	&	0.825	&	0.767	&	0.847	&	0.824	&	0.838	&	0.815	&	0.820	&	0.854	&	0.788	&	0.837	&	0.848	&	0.841	&	0.842	&	\textbf{0.870}	\\
	&	$HCE_\gamma\downarrow$	&	3218	&	3601	&	3654	&	3760	&	3678	&	3653	&	3683	&	3803	&	3806	&	3709	&	3864	&	3999	&	3690	&	3817	&	3819	&	3331	&	\textbf{2888}	\\
	\hline 
	\hline
\multirow{6}{*}{\begin{sideways}\tabincell{c}{\textbf{Overall}\\\textbf{DIS-TE (1-4)}}\end{sideways}}	&	$maxF_\beta\uparrow$	&	0.708	&	0.752	&	0.694	&	0.704	&	0.660	&	0.761	&	0.693	&	0.712	&	0.710	&	0.678	&	0.743	&	0.674	&	0.711	&	0.729	&	0.710	&	0.757	&	\textbf{0.799}	\\
	&	$F^w_\beta\uparrow$	&	0.603	&	0.663	&	0.590	&	0.614	&	0.554	&	0.670	&	0.608	&	0.624	&	0.620	&	0.584	&	0.658	&	0.558	&	0.622	&	0.658	&	0.628	&	0.665	&	\textbf{0.726}	\\
	&	$~M~\downarrow$	&	0.103	&	0.086	&	0.103	&	0.098	&	0.112	&	0.083	&	0.101	&	0.097	&	0.095	&	0.105	&	0.087	&	0.110	&	0.095	&	0.085	&	0.094	&	0.084	&	\textbf{0.070}	\\
	&	$S_{\alpha}\uparrow$	&	0.759	&	0.783	&	0.732	&	0.750	&	0.728	&	0.791	&	0.747	&	0.756	&	0.755	&	0.729	&	0.781	&	0.737	&	0.758	&	0.770	&	0.754	&	0.792	&	\textbf{0.819}	\\
	&	$E_{\phi}^{m}\uparrow$	&	0.798	&	0.835	&	0.797	&	0.819	&	0.776	&	0.835	&	0.822	&	0.827	&	0.819	&	0.810	&	0.840	&	0.778	&	0.825	&	0.850	&	0.832	&	0.834	&	\textbf{0.858}	\\
	&	$HCE_\gamma\downarrow$	&	1202	&	1313	&	1339	&	1401	&	1395	&	1333	&	1411	&	1427	&	1442	&	1365	&	1432	&	1513	&	1359	&	1457	&	1426	&	1218	&	\textbf{1016}	\\
        \hline
	\end{tabular}
	}
\end{table*}

\begin{figure*}[t]
    \centering
    \begin{overpic}[width=\textwidth]{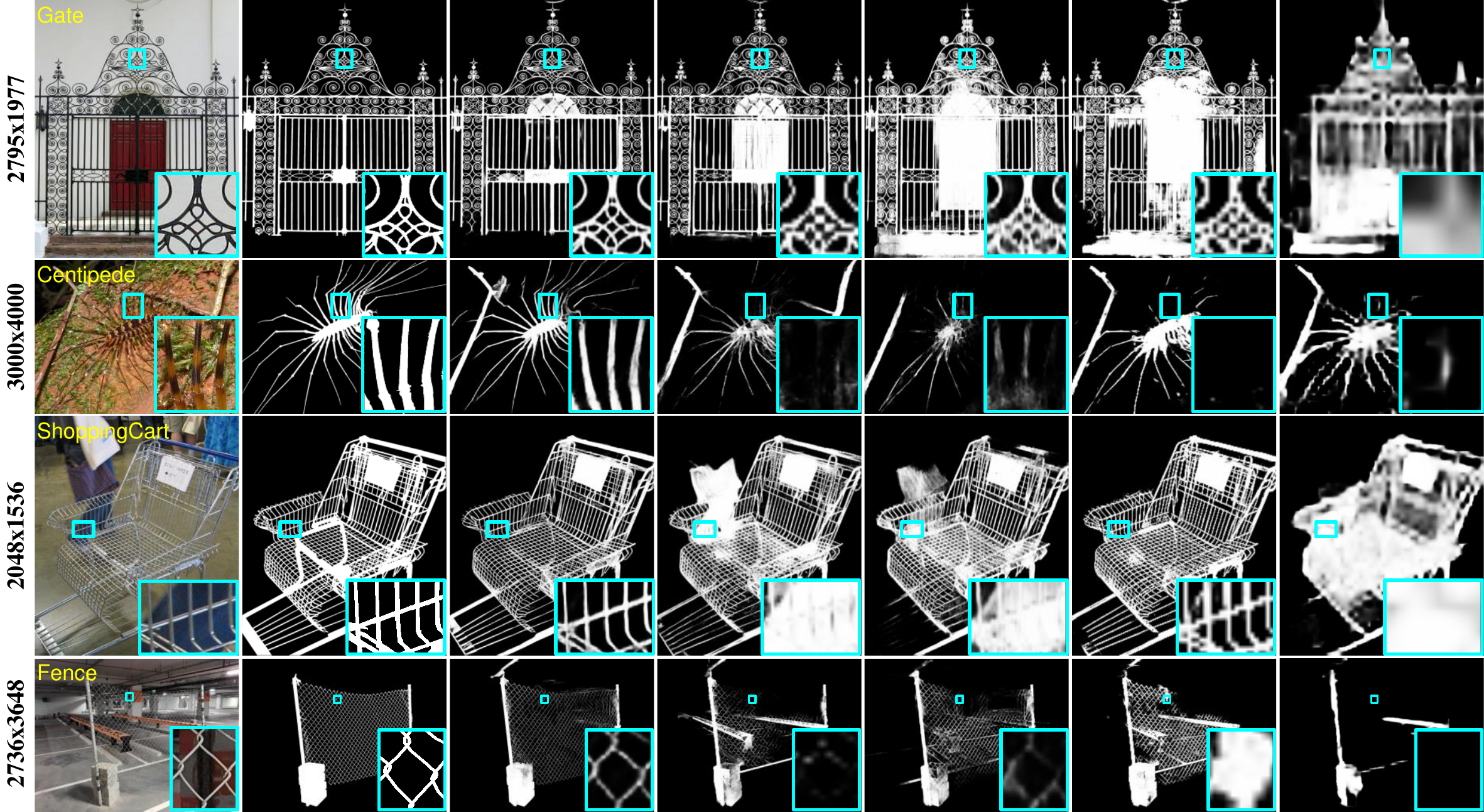}
    \put(6,-2) {\small Image}
    \put(21,-2) {\small GT}
    \put(34,-2) {\small Ours}
    \put(46,-2) {\small U$^2$-Net}
    \put(58,-2) {\small HyperSeg-M}
    \put(75,-2) {\small HRNet}
    \put(88,-2) {\small SINet-V2}
    \end{overpic}
    \caption{\small Qualitative comparisons of \ourmodel~with four baselines. Refer to the \supp{SM} for more results.}\label{fig:qual}
    \vspace{-10pt}
\end{figure*}

\section{\ourdataset~Benchmark}
As discussed above, our DIS5K is built from scratch to cover highly diversified objects with very different geometrical structures and image characteristics. One of the most important reasons is to exclude the existing datasets' possible biases (to specific image or object characteristics). Therefore, its diversities (\eg, resolutions, image characteristics, object complexities, labeling accuracy) and distributions differ from the existing datasets. All models are trained, validated, and tested on DIS-TR, DIS-VD, and DIS-TE, respectively, to provide a fair comparison. Currently, cross-dataset evaluations~\cite{torralba2011unbiased} are not conducted mainly because their labeling accuracy is not consistent with ours.

\noindent\textbf{Metrics.}
To provide relatively comprehensive and unbiased evaluations, six different metrics, including maximal F-measure ($F_\beta^{mx}\uparrow$) \cite{achanta2009frequency}, weighted F-measure ($F^w_\beta\uparrow$) \cite{Margolin2014HowTE}, mean absolute error ($M\downarrow$) \cite{perazzi2012saliency}, structural measure ($S_{\alpha}\uparrow$) \cite{fan2017structure}, mean enhanced alignment measure ($E_{\phi}^{m}\uparrow$) \cite{fan2018enhanced,fan2021cognitive} and our human correction efforts ($HCE\gamma\downarrow$), are used to evaluate the performance from different perspectives. 

\noindent\textbf{Competitors.}
To provide comprehensive evaluations, we compared our \ourmodel~with 16 popular networks designed for different segmentation tasks, including 
(i) popular medical image segmentation model, U-Net \cite{ronneberger2015u}; 
(ii) salient object detection models such as BASNet \cite{qin2019basnet}, GateNet \cite{zhao2020suppress}, F$^3$Net~\cite{wei2020f3net}, GCPA~\cite{chen2020global} and U$^2$-Net \cite{qin2020u2}; 
(iii) models designed for COD like SINet-V2\cite{fan2021concealed} and PFNet \cite{Mei_2021_CVPR}; 
(iv) semantic segmentation models: PSPNet \cite{zhao2017pyramid}, DeepLab-V3+~\cite{chen2018encoder} and HRNet~\cite{WangSCJDZLMTWLX19}; 
(v) real-time semantic segmentation models: BiSeNetV1 \cite{yu2018bisenet}, ICNet \cite{zhao2018icnet}, MobileNet-V3-Large \cite{DBLP:conf/iccv/HowardPALSCWCTC19}, STDC \cite{fan2021rethinking} and HyperSegM \cite{nirkin2020hyperseg}. 
All models are re-trained using DIS-TR set (on Tesla V100 or RTX A6000) and the time costs in \tabref{tab:compSOA} are all tested on RTX A6000.

\subsection{Quantitative Evaluation}
From \tabref{tab:compSOA}, compared with the 16 SOTA models, our \ourmodel ~achieves the most competitive performance across all metrics. 
According to our observations, the performance of different models may be partially related to the model input size and the spatial size of their feature maps. As we know, most of the segmentation models introduce the existing image classification backbones to construct their encoder-decoder architectures for image segmentation tasks. However, some of the backbones like ResNet-50 \cite{he2016deep} starts with an input convolution layer (stride of two) followed by a pooling operation (stride of two) to reduce the spatial size of the feature maps to a quarter of the input size, which leads to the loss of much spatial information and significant performance degradation. When the shape of the to-be-segmented target is close to convex, the information lost and performance degradation is less significant. However, many objects in \ourdataset~are non-convex, and they have very complicated and fine structures. Therefore, \ourdataset~requires the models to keep the spatial information as much as possible, which is challenging to most models. 

\subsection{Qualitative Evaluation}

\figref{fig:qual} presents qualitative comparisons between our approach and four SOTA baselines. Our model achieves promising results on the diverse scenes no matter that they are salient (gate), camouflaged (centipede), thin (shopping cart) or meticulous (fence) objects, demonstrating the generalization capability of our \ourmodel~baseline.

\subsection{Ablation Study}
To validate the effectiveness of our adaptation on recent SOTA model \eg, U$^2$-Net and our newly proposed intermediate supervision strategy, we conduct comprehensive ablation studies. 

\noindent\textbf{Input Size.} As can be seen in \tabref{tab:abl}, 
a larger input size can improve the performance of U$^2$-Net. However, it also increases the GPU memory costs so that we need to reduce the batch size (3 on Tesla V100, 32 GB) when the input size is $1024\times1024$, which degrades the performance. Our simple and effective variant (\ie, Adp, 4$^{rd}$ row) addresses this memory issue and improves the performance.

\noindent\textbf{Supervision on Different Decoder Stages.}
In \tabref{tab:abl}, Last-$S$ means the intermediate supervision is applied on the last $S$ decoder stages. As shown, applying intermediate supervisions on the Last-6 stage gives relatively better performance, which is used as our default setting.

\noindent\textbf{Different Loss.}
The results of different losses show that $L_2$ is better than $KL$ divergence and $L1$. Besides, sharing the ``outconvs'', which transform the deep feature maps to the segmentation probability maps, of the GT encoders and the segmentation decoders leads to negative impacts. 

\begin{table}[t!]
    \centering
    \scriptsize
    \caption{\small Ablation studies on our DIS-VD set.}\label{tab:abl}
    \renewcommand{\arraystretch}{1.0}
    \setlength\tabcolsep{1.0pt}
    \begin{tabular}{l|cccccc}
        \hline 
         Settings & 
         $F_\beta^{mx}\uparrow$ & $F^w_\beta\uparrow$ & $~M~\downarrow$ & $S_{\alpha}\uparrow$ & $E_{\phi}^{m}\uparrow$ & $HCE_\gamma\downarrow$\\
         \hline 
        U$^2$-Net $320^2$ (baseline) &.748 & .656 &.090 &.781 & .823 &1413\\
        U$^2$-Net $512^2$  &.769 &.677 &.085 &.789 &.826 & 1146\\
        U$^2$-Net $1024^2$ &.764 &.667 &.088 &.792 &.820 & 1085\\
        \rowcolor{mygray}
        U$^2$-Net $1024^2$ (\textbf{Adp}) &\textbf{.776} &\textbf{.695} &\textbf{.080} &\textbf{.804} &\textbf{.844} & \textbf{1076}\\
        \hline 
        Adp+Last-1($L_2$) &.777 &.695 &.080 &.799 &.840 &1115\\
        Adp+Last-2($L_2$) &.778 &.704 &.079 &.803 &.847 &\textbf{1049}\\
        Adp+Last-3($L_2$) &.788 &.708 &.079 &\textbf{.812} &.845 &1078\\
        Adp+Last-4($L_2$) &.782 &.703 &.079 &.807 &.849 &1063\\
        Adp+Last-5($L_2$) &.788 &\textbf{.715} &\textbf{.074} &.811 &\textbf{.853} &1059\\
        \rowcolor{mygray}
        Adp+Last-6($L_2$) &\textbf{.790} &.710 &\textbf{.074} &.810 &.852 & 1056\\
        \hline 
        Adp+Last-6($KL$) &.770 &.684 &.084 &.794 &.837&1092\\
        Adp+Last-6($L_1$) &.770 &.686 &.080 &.797 &.837&1144\\
        Adp+Last-6($L_2$) (shared outconv) &.745 &.646 &.094 &.779 &.813 &1191\\
        \hline
        Adp+Last-6($L_2$,sd(1))  &.786 &.706 &.076 &.807 &.844 &1086\\
        Adp+Last-6($L_2$,sd(58)) &.790 &.709 &.078 &.812 &.848 &\textbf{1085}\\
        Adp+Last-6($L_2$,sd(472)) &.790 &.712 &.075 &.812 &.852 &1071\\
        \rowcolor{mygray}
        Adp+Last-6($L_2$,sd(5289)) (\textbf{\ourmodel}) &\textbf{.791} &\textbf{.717} &\textbf{.074} &\textbf{.813} &\textbf{.856} &1116\\
         \hline 
    \end{tabular}
\end{table}

\noindent\textbf{Random Seeds.}
To study the influences of random weights initialization, we trained the same GT encoder multiple times with weights initialized by different random seeds. 
As seen, although the performance produced by different random seeds are different, 
their variations are minor, and all of them are better than that of the models (U$^2$-Net and Adp) trained without our intermediate supervision strategy. Since the model from seed 5289 ranks the 1$^{st}$ on five out of six overall metrics, we use this model as our \ourmodel.

\section{Conclusions}
We have systematically studied the highly accurate dichotomous image segmentation (DIS) task from both the application and the research perspective. To prove that the task is solvable, we have built a new challenging \textbf{\ourdataset}~dataset, introduced a simple and effective intermediate supervision network, called \ourmodel, to achieve high-quality segmentation results in real-time, and designed a novel Human Correction Efforts (\textbf{HCE}) metric by considering the shape complexities for applications.
With an extensive ablation study and comprehensive benchmarking, we obtained that our newly formulated DIS task is solvable.

\noindent\textbf{Broader impacts.} 
This work may greatly facilitate the applications of segmentation techniques in both academia and industry, and hereby invite all the researchers in related fields to collaborate and improve the whole eco-system.

{\small
\bibliographystyle{ieee}
\bibliography{main_supp_cvpr2022}
}

\newpage 

\twocolumn[{%
\renewcommand\twocolumn[1][]{#1}%
\begin{center}
    \vspace{-20pt}
    \part{\large \bf{Highly Accurate Dichotomous Image Segmentation}}
  \vspace{-5pt}
  {\large \textit{-Supplementary Material}} 
\end{center}
\maketitle
\vspace{-20pt}
\begin{center}
    \centering
    \captionsetup{type=figure}
    \includegraphics[width=0.9\textwidth,height=1.143\textwidth]{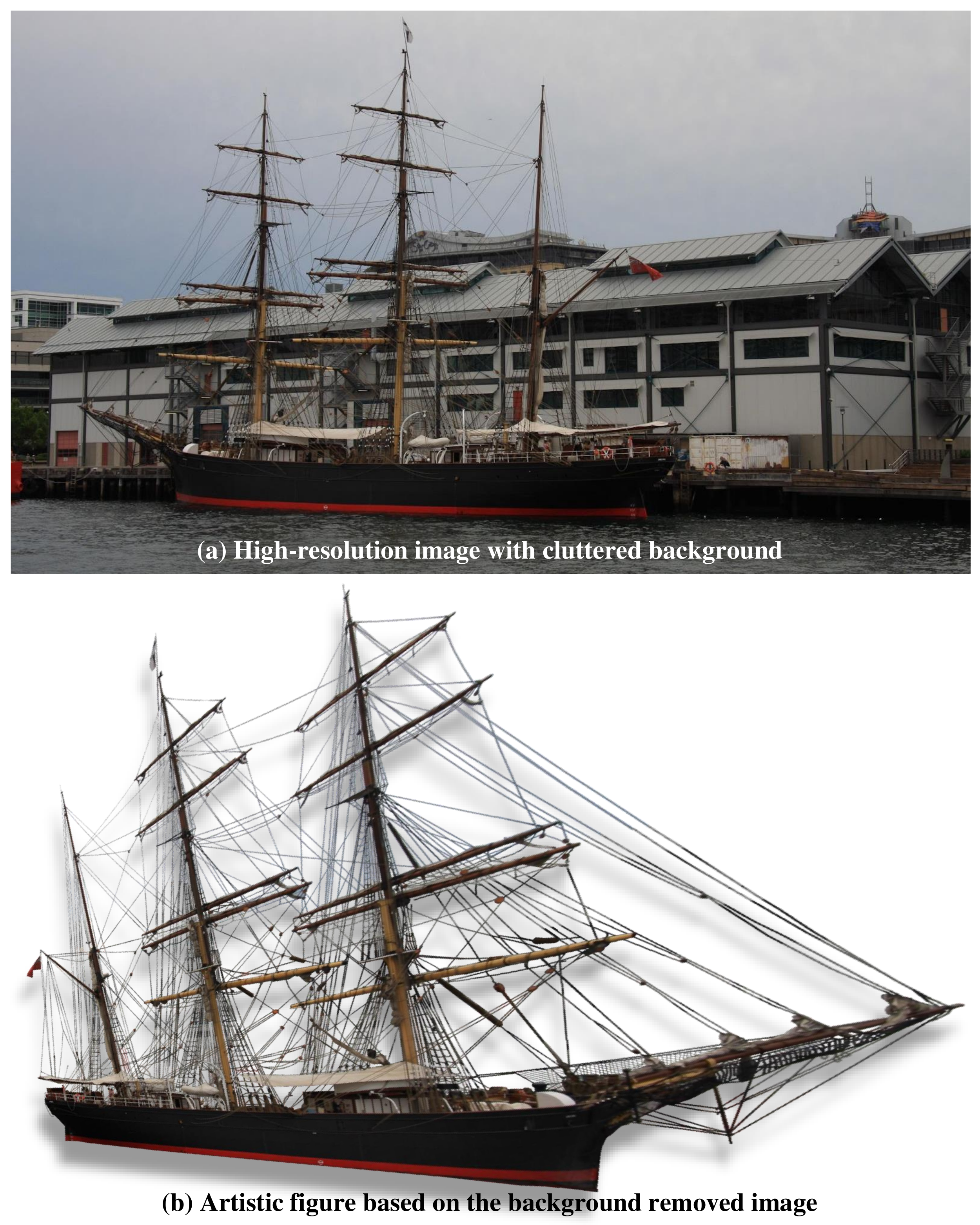}
    \vspace{-10pt}
    \captionof{figure}{\small Demo application: artistic figure generated based on a sample of our \ourdataset~dataset.
    }
    \label{fig:intro}
\end{center}%
}]


\section{Related Work}

\subsection{Multi-class \textit{vs.} Dichotomous Segmentation}
Multi-class (\eg semantic \cite{long2015fully}, panoptic \cite{li2021fully}) segmentation aims at simultaneously labeling all the pixels in an image of complex scenario \cite{zhou2017scene,Cordts2016Cityscapes}, which contains many different objects, with the pre-defined multiple categories encoded in one-hot vectors. However, the one-hot representation of the categories is memory exhaustive when the number of categories is huge (\eg, 10,000 categories), especially on high-resolution images. Besides, some input images only contain objects from several categories (\eg, one or two). Outputting the full-length one-hot dense predictions (10,000 categories) is not a resource-saving option. 
A possible alternative could be a two-step solution: ``detection + segmentation'', in which a bounding box and category of the certain object can be predicted first. The segmentation process can then be conducted in a dichotomous way within the bounding box region by producing a single-channel probability map (\eg, similar to Mask R-CNN\cite{he2017mask}. However, Mask R-CNN still uses the one-hot representation in the segmentation step).

Moreover, many practical applications, such as image editing, art design, shape from silhouette, robot manipulation, are usually category-agnostic, where the applications require highly accurate segmentation results of certain objects regardless of their categories. Different from the images of complex scenarios in semantic \cite{lin2014microsoft} or panoptic \cite{zhou2017scene} segmentation, the images in these applications usually contain one or a few objects with very high resolutions, less occlusions. 
To this end, many related tasks have been proposed, such as salient object detection (SOD) \cite{movahedi2010design,yan2013hierarchical,DBLP:journals/pami/LiuYSWZTS11,yang2013saliency,DBLP:journals/pami/ChengMHTH15,wang2017learning,wang2019salient}, salient object in clutter (SOC) \cite{fan2018SOC}, high-resolution salient object detection (HRS) \cite{zeng2019towards}, camouflaged object detection (COD) \cite{le2019anabranch, chameleon,fan2020camouflaged}, thin object segmentation (TOS) \cite{liew2021deep}, meticulous object segmentation (MOS) \cite{yang2020meticulous}, video object segmentation (VOS) \cite{perazzi2016benchmark}, class-agnostic very high-resolution segmentation (VHRS) \cite{cheng2020cascadepsp}, \etc. 
Most of these tasks try to solve dichotomous segmentation problems on images which are sharing specific characteristics. 
The exclusive mechanisms for certain tasks are barely used so that their problem formulations are almost the same, which means most of these tasks are data-dependent. Simply combining these tasks by merging their datasets is not a decent option because these tasks' image resolutions and labeling qualities are diversified.  

Considering these facts, we re-formulate a new category-agnostic dichotomous segmentation task, \textit{highly accurate Dichotomous Image Segmentation (DIS)}, where achieving highly accurate segmentation results of objects with diversified shapes and structures is the key concern. 

\subsection{Datasets} 

Datasets are the basis of most computer vision tasks. 
In the past decades, many segmentation datasets for related tasks have been created. For example, semantic (PASCAL-VOC\cite{everingham2010pascal}, MS-COCO\cite{lin2014microsoft}) and panoptic (Cityscapes\cite{Cordts2016Cityscapes}, ADE20K \cite{zhou2017scene}) segmentation (SMS) datasets usually contain large number of images with multiple objects from different categories in each of them. 
But they either have low geometrical labeling accuracy or relatively small resolutions, where details of objects are hard to be included and segmented. 
The entity segmentation (ES) \cite{qi2021open} datasets proposed for class-agnostic segmentation has similar issues. Images in the salient object detection (SOD) \cite{movahedi2010design, li2014secrets, yang2013saliency, DBLP:journals/pami/ChengMHTH15, wang2017learning} and camouflaged object detection (COD) \cite{fan2020camouflaged} datasets are usually low-resolution ones, which contains objects with simple structures. The high-resolution salient object detection (HRS) \cite{zeng2019towards,perazzi2016benchmark} datasets have higher resolution, but they are built upon images with objects of simple structures similar to that in SOD and COD datasets. The meticulous object segmentation (MOS) \cite{yang2020meticulous} and thin object segmentation (TOS) \cite{liew2021deep} datasets show competitive resolution and object structure complexity characteristics. However, MOS is too small to enable thorough training and comprehensive evaluation, while the TOS dataset is built with synthetic images. 
Therefore, there is a need for a new \textit{extendable} \textit{large-scale} dataset built upon the \textit{high-resolution} images with \textit{diversified object structure complexities} and \textit{highly accurate labeling}.

\subsection{Existing Models} 
Models are the cores of vision tasks. 
Currently, deep models are the most popular solutions for most of the segmentation tasks. 
Many different deep architectures have been proposed to achieve better performance, such as FCN-based \cite{long2015fully} feature aggregation models \cite{DBLP:conf/iccv/ZhangWLWY17, luo2017non, hou2017deeply, DBLP:conf/ijcai/ZhangLLS18, DBLP:conf/eccv/ChenTWH18, WangSCJDZLMTWLX19,zhao2020suppress,wei2020f3net}, Encoder-Decoder architectures \cite{badrinarayanan2017segnet,ronneberger2015u,qin2020u2,chen2020global}, 
Coarse-to-Fine (or Predict-Refine) models \cite{DBLP:conf/iccv/WangBZZL17, deng2018r3net, wang2018detect,qin2019basnet,cheng2020cascadepsp,HRRN_ICCV2021,DBLP:journals/corr/abs-2108-11515}, 
Vision Transformers \cite{zheng2020rethinking,liu2021visual}, \etc. 
Besides, many real-time models \cite{zhao2018icnet,yu2018bisenet,li2019dfanet,orsic2019defense,hu2020temporally,Fan_2021_CVPR,nirkin2020hyperseg} are developed to balance the performance and time costs. To achieve highly accurate results in our DIS, the models are expected to capture fine details (and complicated structures) and large components of the diversified objects from large-size (\eg, 2K, 4K or even larger) images with affordable memory, computation and time costs. These requirements are very challenging to the existing segmentation models. 
Therefore, more effective, more efficient, and more stable models are needed. 

\subsection{Over-fitting \textit{vs.} Regularization} 

Most deep segmentation models can fit the training sets very well (training accuracy close to $100\%$) while having different performances on the testing sets. 
To the best of our knowledge, there could be two main reasons. 
On one hand, the ``distributions'' between the training, validation, and testing sets are not guaranteed to be the same, which leads to performance degradation of almost all the models on testing sets. 
On the other hand, different model architectures have diversified capabilities of feature representations, which means they are more likely to fit the training sets in very different ways, namely, transforming the input images into other high-dimensional spaces. 
Most of the works are following this direction to develop more representative architectures. 
However, there lacks an effective way to measure the representation capabilities of these architectures before testing, so the model design is usually conducted by trial and error. 
Hence, some researchers turn to search for different ways for reducing over-fitting. Different supervision strategies, such as weights regularization \cite{Goodfellow-et-al-2016}, dropout \cite{DBLP:journals/jmlr/SrivastavaHKSS14}, dense (deep) supervision \cite{lee2015deeply,qin2020u2,xie2015holistically}, hybrid loss \cite{luc2016semantic, qin2019basnet,zhao2019egnet} and so on, have been proposed. 
The dense (deep) supervision \cite{lee2015deeply,xie2015holistically,qin2020u2}, which imposes ground truth supervisions on the side outputs from several of the deep intermediate layers, is one of the most popular ways. 
However, transforming the deep intermediate features (multi-channel) into the side outputs (single-channel) in dichotomous image segmentation (DIS) is essentially a dimension reduction operation, which leads to information losses, so that weaken the supervisions.
In this paper, instead of developing more complicated deep architectures, we follow the dense supervision idea but develop a simple yet more effective supervision strategy, \textbf{intermediate supervision}, to directly enforce the supervisions on high-dimensional intermediate deep features in addition to the side outputs. 

\subsection{Evaluation Metrics}

The evaluation strategies and metrics are expected to provide comprehensive and practically meaningful evaluations to analyze the prediction qualities. 
Currently, many evaluation metrics, such as IoU, boundary IoU \cite{DBLP:conf/cvpr/ChengGDBK21}, F-measure \cite{achanta2009frequency}, boundary F-measure \cite{ehrig2005relaxed, qin2019basnet}, boundary displacement error (BDE) \cite{DBLP:conf/eccv/FreixenetMRMC02}, boundary IoU \cite{DBLP:conf/cvpr/ChengGDBK21}, structural measure ($S_m$) \cite{fan2017structure}, Mean Absolute Error (MAE) \cite{perazzi2012saliency}, and so on, are usually defined based on consistencies (or inconsistencies) between the model predictions and the ground truth. 
Most of them are usually biased to certain types of structures. 
For example, IoU and F-measure mainly rely on the object components with large areas while neglecting the fine details with relatively small areas. 
To alleviate this issue, boundary F-measure, BDE, and boundary IoU are developed to focus on the boundary quality. 
However, these boundary-based metrics are often highly dependent on those long smooth boundary segments' qualities while failing to describe the qualities of those short jagged boundary segments. 
Besides, the above metrics are mostly defined from the mathematical or cognitive perspective; none of them are able to reflect the barriers (or costs) of applying the predictions in real-world applications, where certain accuracy requirements have to be satisfied. 
To address these issues, we propose a novel metric, named as human correction efforts (HCE), to measure the barriers by approximating the human efforts for correcting the faulty regions of the model predictions. 

\begin{figure*}[thbp]
    \centering
    \includegraphics[width=0.96\textwidth]{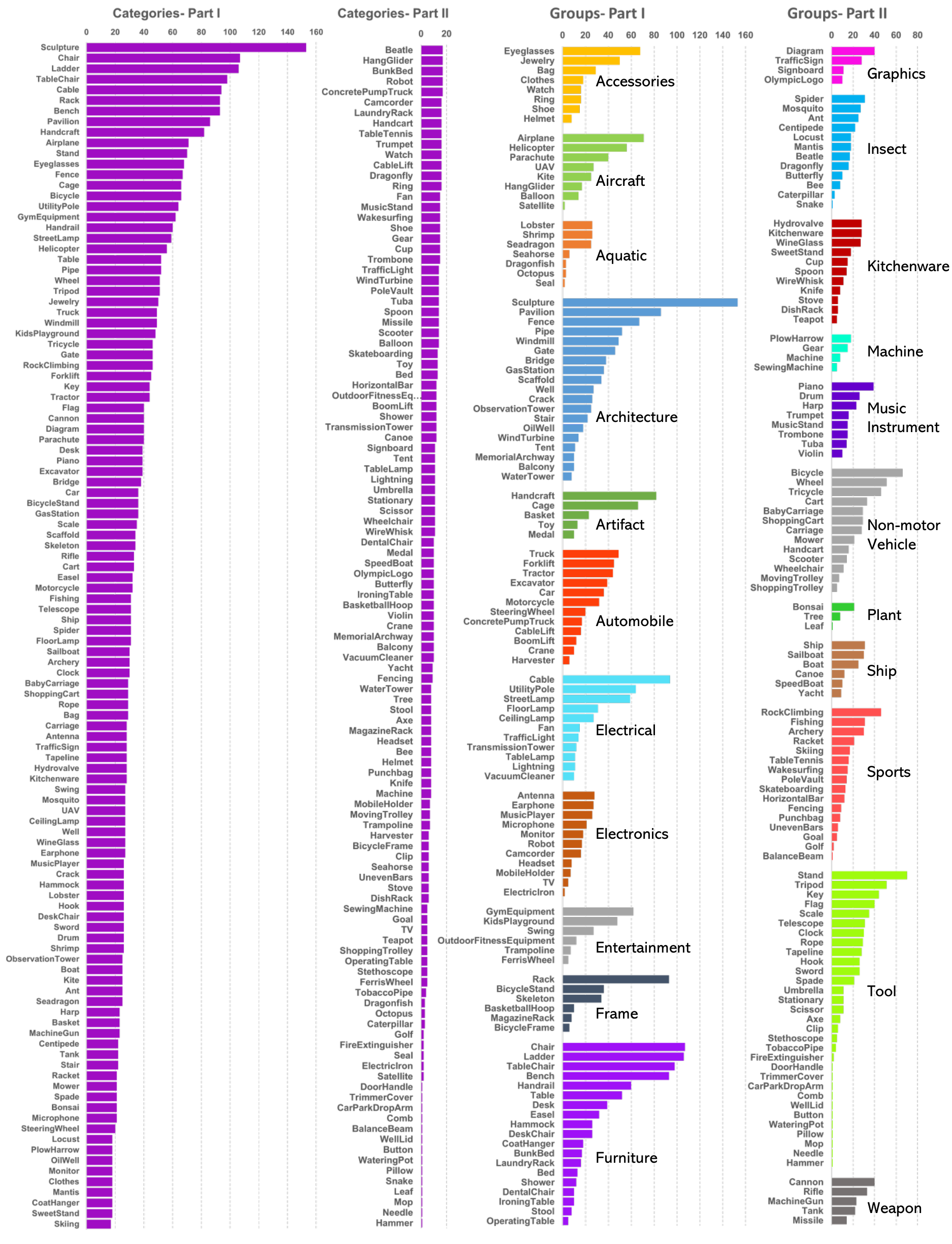}
    \caption{\small Number of images per-category and per-group.}
    \label{fig:DIS-hist}
\end{figure*}

\section{More Details of DIS5K Dataset}
\subsection{Per-category and per-group statistics}
Fig.~\ref{fig:DIS-hist} illustrates the number of images per-category and per-group. Our DIS5K contains 5,470 images from 225 categories divided into 22 groups. The average numbers of images per category and per group are around 24 and 249, respectively.

\begin{figure*}[thbp]
    \centering
    \includegraphics[width=0.96\textwidth]{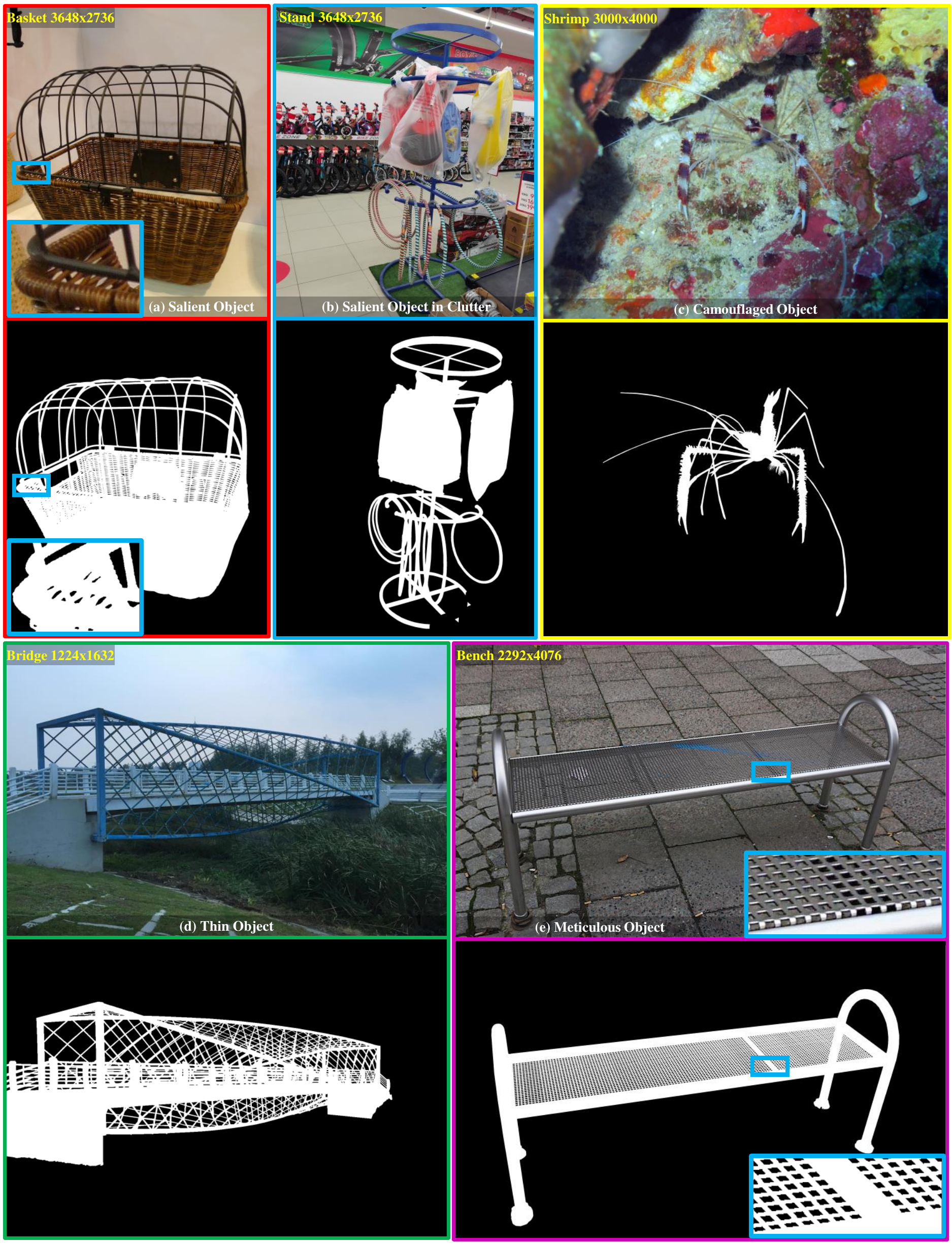}
    \caption{\small Sample images and ground truth masks with objects of certain characteristics.}
    \label{fig:DIS-tasks}
\end{figure*}

\subsection{Typical Samples from DIS5K} 
Fig.~\ref{fig:DIS-tasks} shows some samples from our DIS5K, which have certain characteristics similar to that of the existing dichotomous segmentation tasks, such as salient object detection (SOD) \cite{wang2017learning}, salient object in clutter (SOC) \cite{fan2018SOC}, camouflaged object detection (COD) \cite{fan2020camouflaged}, thin object segmentation (TOS) \cite{liew2021deep}, meticulous object segmentation (MOS) \cite{yang2020meticulous}. It is worth mentioning that ``salient object'', ``salient object in clutter'' and ``camouflaged object'' are mainly defined based on the contrast between foreground targets and background environments. In comparison, ``thin object'' and ``meticulous object'' are based on the geometric structure complexities of the foreground targets. Therefore, the first three types of objects and the last two types of targets are not exclusive. For example,  the basket in Fig.~\ref{fig:DIS-tasks} (a) and the shrimp in Fig.~\ref{fig:DIS-tasks} (c) can also be taken as meticulous because the basket has many holes and the shrimp has jagged boundaries. Besides, the boundaries among SOD, SOC, and COD and the boundaries between TOS and MOS are blurring. There are some overlaps between them in terms of data samples. Our DIS5K contains all the above types of images paired with highly-accurate ground truth masks. 

\subsection{Object Structure Analysis} 
\begin{figure*}[thbp]
    \centering
    \includegraphics[width=\textwidth]{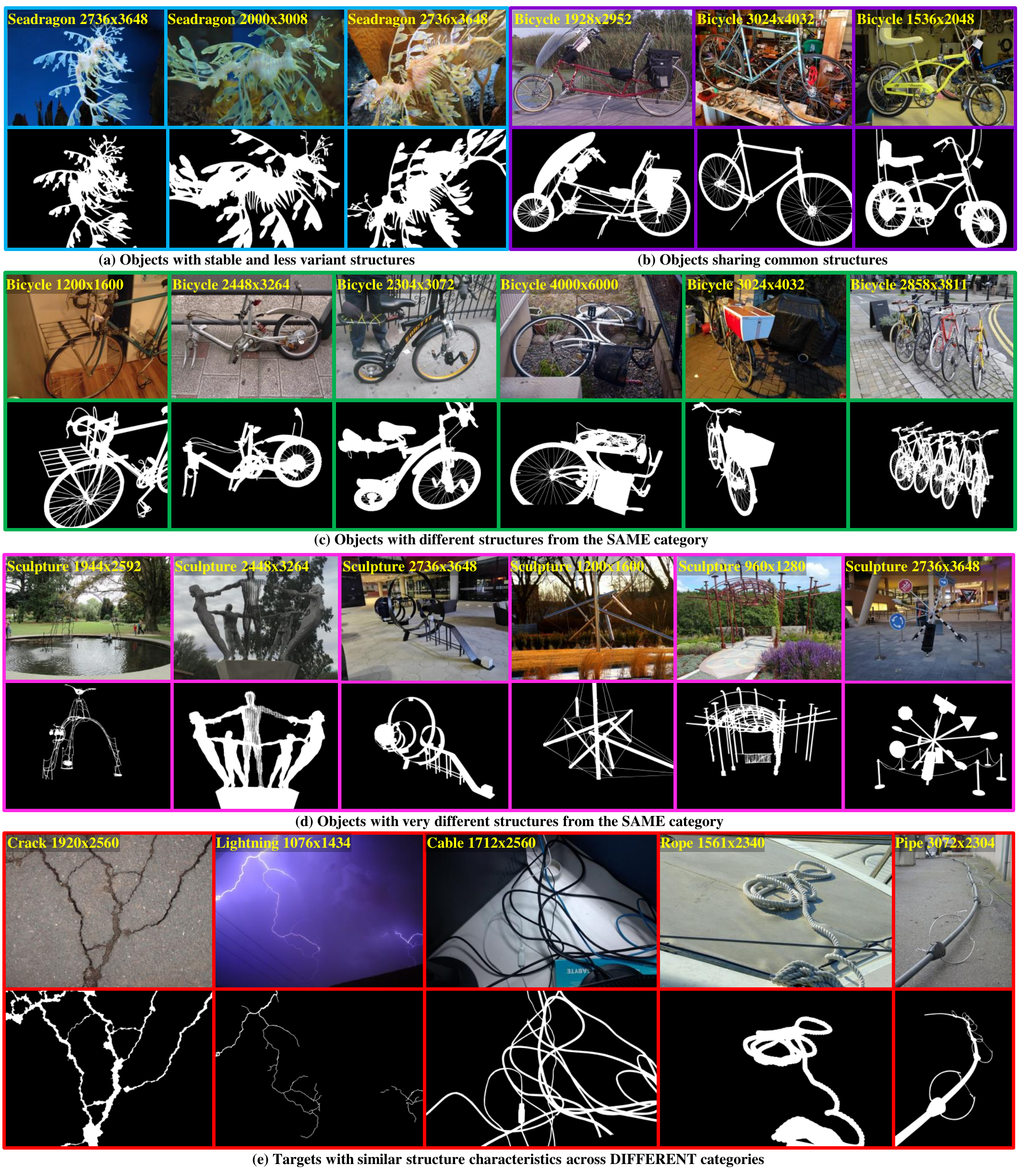}
    \caption{\small Structure analysis of inter- and intra-category targets.}
    \label{fig:DIS-structure}
\end{figure*}

In addition to the above mentioned image characteristics, there are also some interesting observations on object structures from our DIS5K, as shown in Fig.~\ref{fig:DIS-structure}.

\begin{table*}[t]
    \centering
    \caption{\small
    Image dimension and object complexity of the subsets of \ourdataset. 
    $\sigma_{(\cdot)}$ is the standard deviation of the corresponding index. 
    }\label{tab:te14}
    \renewcommand{\arraystretch}{1.5}
    \setlength\tabcolsep{6.0pt}
    \resizebox{\textwidth}{!}{
    \begin{tabular}{r|l||r|rrr|rrr}
        \toprule
        \multirow{2}{*}{Task} & \multirow{2}{*}{Dataset} & Number & \multicolumn{3}{c|}{Image Dimension} & \multicolumn{3}{c}{Object Complexity}\\
        \cmidrule(lr){3-3}
        \cmidrule(lr){4-6} \cmidrule(lr){7-9}
         &  & $I_{num}$ & $H\pm\sigma_H$ & $W\pm\sigma_W$ & $D\pm\sigma_D$ & $IPQ\pm\sigma_{IPQ}$ & $C_{num}\pm\sigma_C$ & $P_{num}\pm\sigma_P$\\
        \hline
        \multirow{6}{*}{DIS} & \textbf{\ourdataset}  &  5470  &  2513.37 $\pm$ 1053.40  &  3111.44 $\pm$ 1359.51  &  4041.93 $\pm$ 1618.26  & 107.60 $\pm$ 320.69  &  106.84 $\pm$ 436.88 & 1427.82 $\pm$ 3326.72 \\ 
        \cmidrule(lr){2-9} 
        & DIS-TR  &  3000  &  2514.15 $\pm$ 1052.45  &  3091.23 $\pm$ 1356.92  &  4028.09 $\pm$ 1612.45  &  69.32 $\pm$ 261.98  &  73.99 $\pm$ 367.81  &  1153.05 $\pm$ 2893.36\\

        & DIS-VD  &  470  &  2472.59 $\pm$ 963.43  &  3102.85 $\pm$ 1308.72  &  4006.49 $\pm$ 1526.56  &  156.85 $\pm$ 349.75  &  163.91 $\pm$ 650.42  &  1954.73 $\pm$ 5119.89\\

        & DIS-TE1  &  500  &  2240.35 $\pm$ 1092.92  &  2678.50 $\pm$ 1291.11  &  3535.32 $\pm$ 1598.89  &  27.13 $\pm$ 29.07  &  6.94 $\pm$ 6.37  &  237.48 $\pm$ 96.27\\

        & DIS-TE2  &  500  &  2402.09 $\pm$ 1047.89  &  3032.25 $\pm$ 1298.45  &  3904.03 $\pm$ 1583.39  &  50.79 $\pm$ 69.85  &  21.20 $\pm$ 16.30  &  583.04 $\pm$ 120.90\\

        & DIS-TE3  &  500  &  2597.15 $\pm$ 988.88  &  3336.51 $\pm$ 1339.10  &  4263.78 $\pm$ 1571.21  &  92.68 $\pm$ 118.99  &  60.96 $\pm$ 40.32  &  1190.93 $\pm$ 255.00\\

        & DIS-TE4  &  500  &  2847.55 $\pm$ 1069.37  &  3527.81 $\pm$ 1412.89  &  4580.93 $\pm$ 1645.86  &  443.32 $\pm$ 667.01  &  482.98 $\pm$ 843.50  &  4858.80 $\pm$ 5618.87\\
        \bottomrule  
    \end{tabular}
    }
\end{table*}

\noindent 
\textbf{Intra-category structure similarity.}
As shown in Fig.~\ref{fig:DIS-structure} (a) and (b), the objects in the same categories are usually showing the same or similar structures and shapes. We call this \textit{intra-category structure similarity}, which is one of the main cues for categorizing. However, the intra-category structure similarity is not always guaranteed. Fig.~\ref{fig:DIS-structure} (c) and (d) show two typical examples against that in different magnitudes. Fig.~\ref{fig:DIS-structure} (c) illustrates some bicycles with variant structures. 
Their differences are mainly caused by components absence (out-of-view imaging, incomplete architecture), variations on the design, view angle changes, co-existence of multiple targets, etc. 
Although the structures of these bicycles are different, they are still sharing some common features, such as wheels, frames, \etc. 
However, objects in some other categories may share no structure similarities. 
For example, the sculptures in Fig.~\ref{fig:DIS-structure} (d) show very different structures and shapes, which indicates low intra-category similarity. 
Because artists or designers usually prefer to design unique architectures, which leads to very diversified object appearances and structures. 
Besides, compared against the relatively stable shapes and structures of the natural targets (\eg, animals, plants), the structures of these human-created objects, which play vital roles in the human-environment interaction of our daily lives, are updated very fast, which further magnifies the challenges in the DIS task. These intra-category dissimilarities significantly increase the difficulty of accurate segmentation and lead to robustness risks. 

\noindent 
\textbf{Inter-category structure similarity.}
In contrary to the low intra-category similarity, there also exist some categories that have high \textit{inter-category structure similarity}. 
Fig.~\ref{fig:DIS-structure} (e) shows some targets from different categories, such as \textit{crack}, \textit{lightning}, \textit{cable}, \textit{rope}, \textit{pipe} and so on. 
These targets are mainly comprised of thin and elongated components. 
For example, the shapes of the crack and the lightning are very close to each other so that they are hard to be differentiated without showing the RGB images. 
The cable, rope, and pipe are also comprised of thin and elongated components with relatively smoother boundaries. 
Besides other targets like roads and rivers in satellite images, vessels in medical images also have similar structural characteristics to those mentioned above. The \textit{inter-category structure similarities} haven't been thoroughly studied, which could be promising directions for exploring the models' explain-abilities and data augmentation strategies. 

Our DIS5K dataset provides relatively richer samples for studying the \textit{intra-category} and \textit{inter-category} similarities and dissimilarities. 
More qualitative and quantitative studies will be helpful to diversified vision tasks, such as image (shape) classification, segmentation, \etc. 
 

\subsection{Attributes of Subsets in DIS5K} 
\begin{figure*}[thbp]
    \centering
    \includegraphics[width=0.96\textwidth]{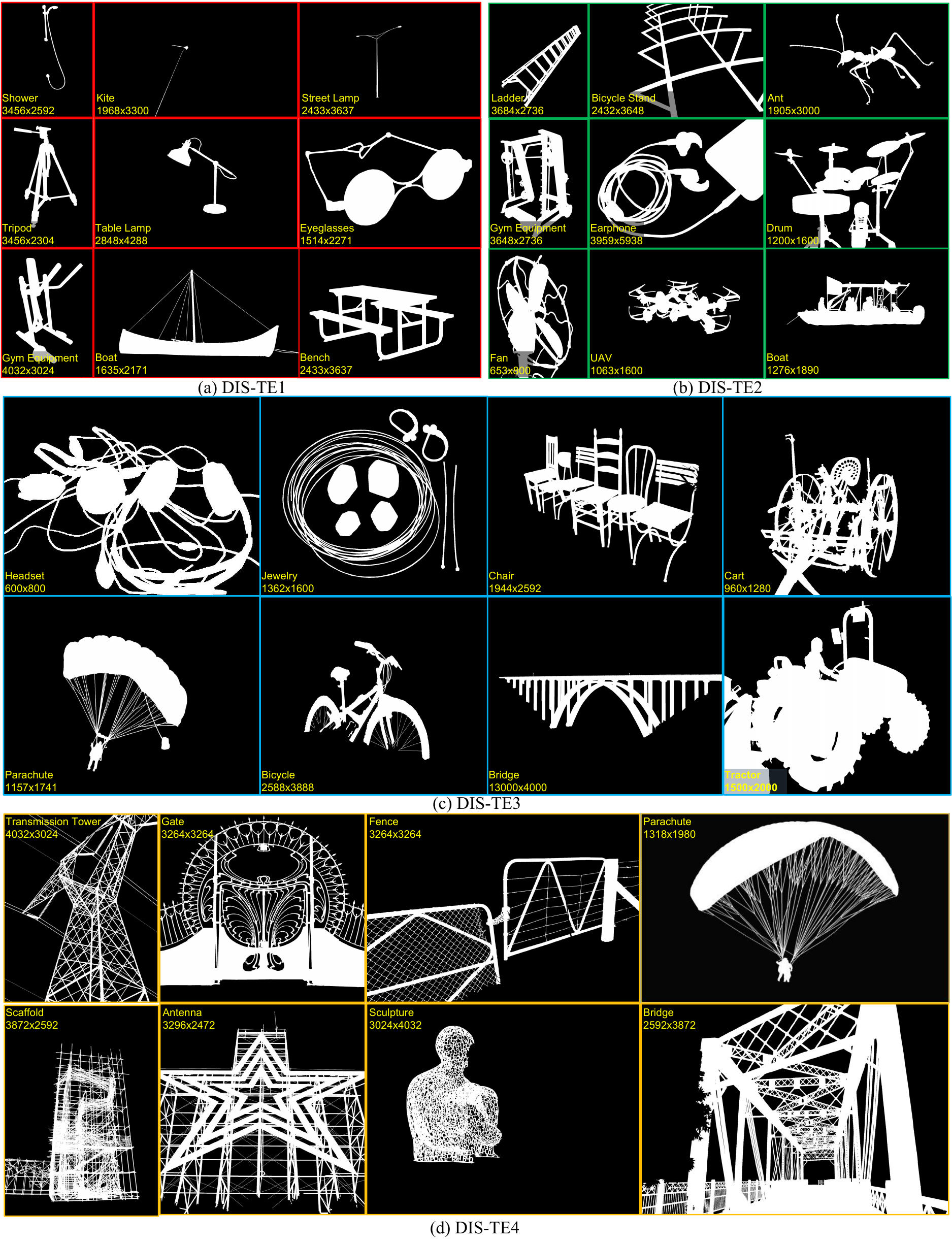}
    \caption{\small Sample ground truth (GT) masks from DIS-TE1, DIS-TE2, DIS-TE3, and DIS-TE4.}
    \label{fig:DIS-te14}
\end{figure*}

Table~\ref{tab:te14} illustrates the essential attributes of the subsets of our DIS5K dataset. As seen, the image dimensions of these subsets are close to each other. At the same time, the complexities of the four testing subsets are in ascending order.  Fig.~\ref{fig:DIS-te14} shows the qualitative comparisons of the structural complexities of our four testing subsets, DIS-T1$\sim$DIS-TE4. Their structure complexities in ascending order can be visually perceived. 







\begin{figure*}[thbp]
    \centering
    \begin{overpic}[width=0.98\textwidth]{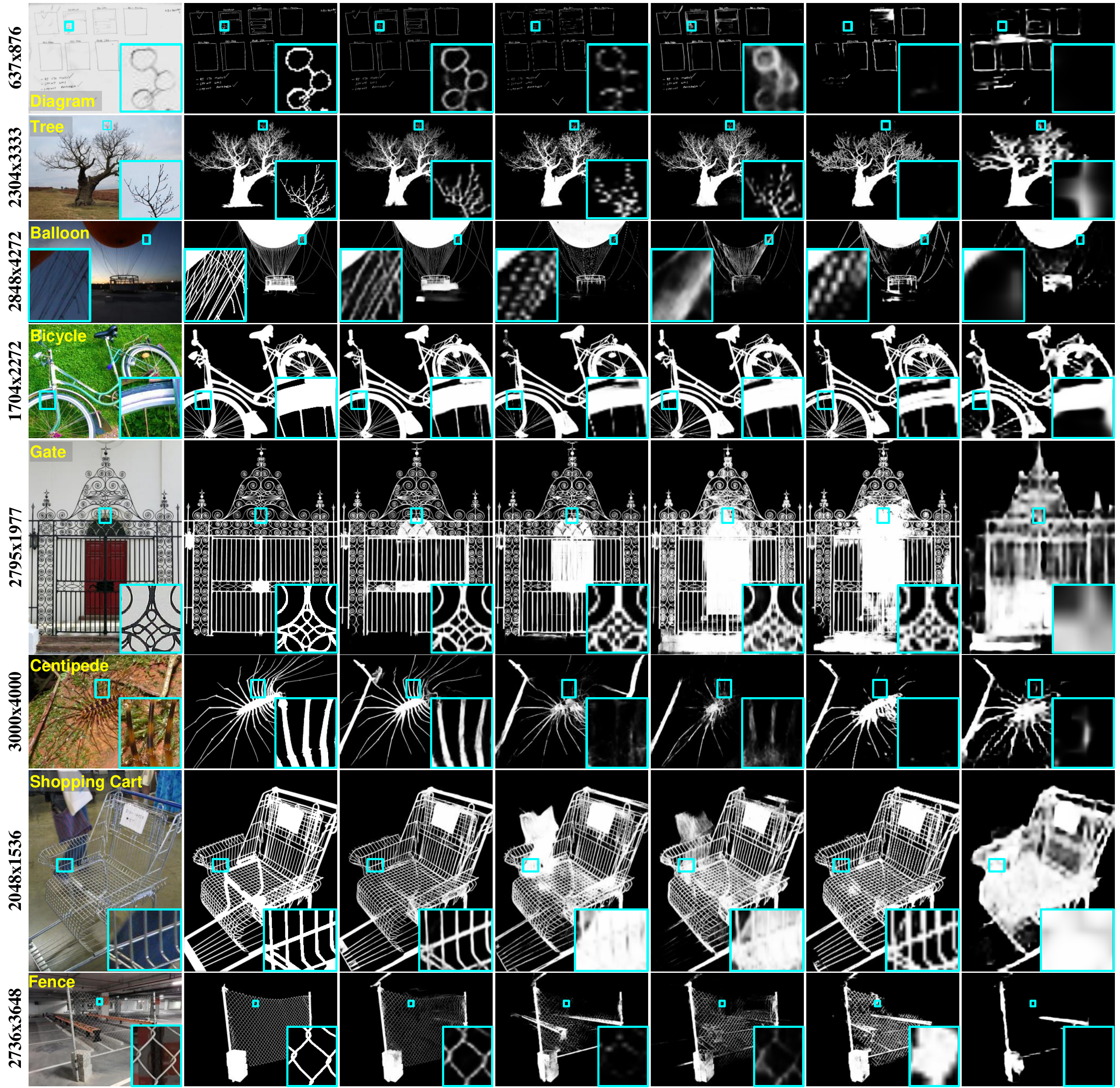}
    \put(8,-1.5) {\small Image}
    \put(23,-1.5) {\small GT}
    \put(36,-1.5) {\small Ours}
    \put(48,-1.5) {\small U$^2$-Net}
    \put(60,-1.5) {\small HyperSeg-M}
    \put(77,-1.5) {\small HRNet}
    \put(90,-1.5) {\small SINet-V2}
    \end{overpic}
    \caption{\small Qualitative comparisons of our model and four cutting-edge baselines.}\label{fig:supp-qual}
\end{figure*}

\section{More Details of Experiments}
\subsection{Implementation details} 
Our models and other baseline models are trained with our DIS-TR (3,000 images) and validated on DIS-VD (470 images). 
The input size of our model is set to $1024\times1024$. It is worth noting that there are many large-size images in our dataset so that the image loading operations in the training and validation are very time-consuming. To address this issue and boost the speed of training and validation, we resize all the input images and their corresponding ground truth to $1024\times1024$ off-line and store them as Pytorch tensor files on the hard disk drive. Although this strategy requires relatively more storage space, it dramatically reduces the time costs for the data loading process in the training and validation stages. 
Our training process consists of two training stages: (i) the training stage of the ground truth encoder and (ii) the training stage of the image segmentation component. 
In both training stages, these three-channel inputs (GT masks are repeated to have three channels) are normalized to [-0.5, 0.5] and only augmented with horizontal flipping. The models weights are initialized by Xavier \cite{DBLP:journals/jmlr/GlorotB10} and optimized with Adam \cite{kingma2014adam} optimizer with the default settings (initial learning rate lr=1e-3, betas=(0.9, 0.999), eps=1e-8, weight decay=0) for both the ground truth encoder and the segmentation component. The batch size of each training step is set to eight, and the validation on DIS-VD is conducted every 1,000 iterations. If the validation results (in terms of $maxF$ and $M$) are improved, the hard disk drive saves the model weights. It is worth mentioning that the loss weights of the dense supervision in the ground truth encoder training and intermediate supervision of the segmentation component training are all set to 1.0. 

According to our experiments, the training process of our ground truth encoder is easy to converge, and it usually takes only 1,000 iterations (stop training when the valid $maxF$ is greater than $0.99$). While the segmentation component of our model usually converges after around 100k iterations, and the whole training process takes less than 48 hours. Besides, all the models are implemented using Pytorch 1.8.0. Some experiments are conducted on a desktop that has a 2.9GHz CPU (128 cores AMD Ryzen Threadripper 3990X), 256 GB RAM and a NVIDIA RTX A6000 GPU. Some other models are trained on NVIDIA TESLA V100 GPU (32 GB).

\subsection{More Analysis of the Experimental Results}

\begin{figure*}[thbp]
    \centering
    \includegraphics[width=\linewidth]{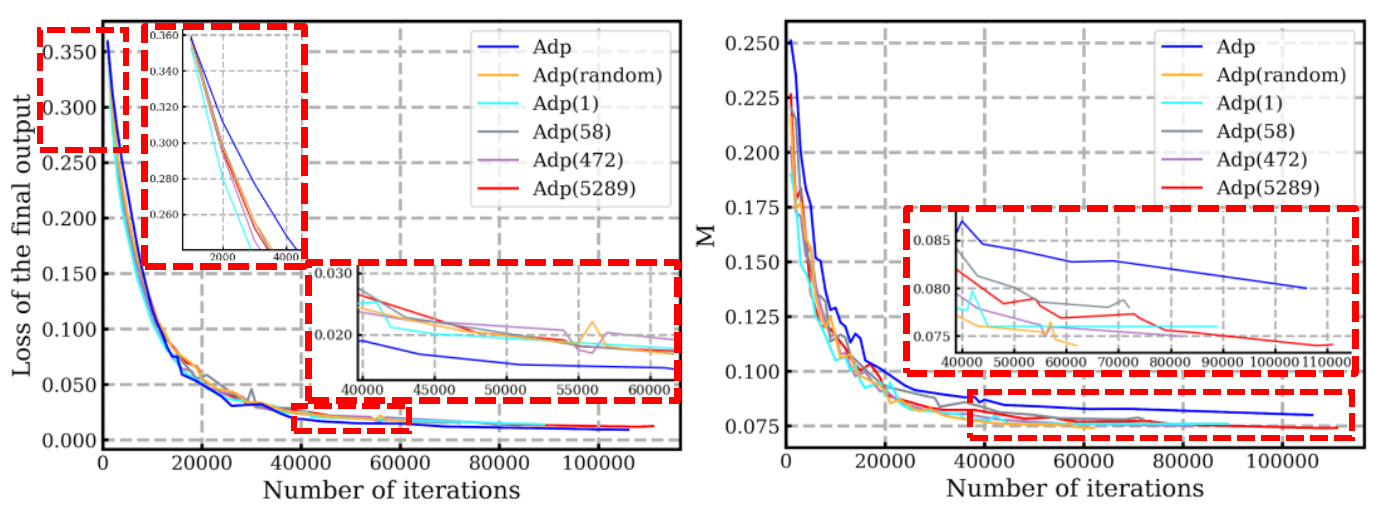}
    \caption{\small Curves of the training loss computed on the last prediction probability map and the Mean Absolute Error ($M$) on our validation set (DIS-VD).}
    \label{fig:tm_curves}
\end{figure*}

\noindent 
\textbf{Performance comparisons among different models.} 
As shown in Table~\ref{tab:compSOA}, our model achieves the most competitive performance against other existing models in terms of almost all the evaluation metrics on different datasets. 
Among the dichotomous segmentation models, U-Net \cite{ronneberger2015u}, BASNet \cite{qin2019basnet}, U$^2$-Net \cite{qin2020u2} and PFNet \cite{Mei_2021_CVPR} performs relatively better against other SOD and COD models. 
Among the semantic segmentation and real-time semantic segmentation models, the results of HRNet \cite{WangSCJDZLMTWLX19} and HyperSeg-M \cite{nirkin2020hyperseg} show more competitive performance. 
Among all the existing models, the performance of HyperSeg-M and U$^2$-Net are close and perform better than other models in both validation and testing sets. Although HRNet and BASNet show slightly inferior performance against HyperSeg-M and U$^2$-Net, they are still more competitive than others. 
Fig.~\ref{fig:supp-qual} provides the qualitative comparisons of our model and other four competitive baseline models. As can be seen, our model achieves the best overall performance on different objects. Surprisingly, other models like U$^2$-Net, HyperSeg-M, and HRNet also obtain encouraging results on certain targets, such as the \textit{tree}, the \textit{gate} and the \textit{shopping cart}, after training on our DIS-TR dataset, which further proves the value of DIS5K. 

\begin{table*}[thbp]
\scriptsize
\centering
	\caption{\small PART-I: Quantitative evaluation on our validation, DIS-VD, and test sets, DIS-TE(1-4), based on groups. ResNet18=R-18. ResNet34=R-34. ResNet50=R-50. Res2Net50=R2-50. DeepLab-V3+=DLV3+. BiseNetV1=BSV1. STDC813=S-813. EffiNetB1=E-B1. MobileNetV3-Large=MBV3. HyperSeg-M=HySM.
	}\label{tab:compSOA-1}
    \renewcommand{\arraystretch}{1.09}
    \setlength\tabcolsep{1.7pt}
	\begin{tabular}{c|r|cccccc|cc|ccc|ccccc|c}
		\hline
        Dataset	&	Metric	&	\tabincell{c}{UNet\\\cite{ronneberger2015u}}	&	\tabincell{c}{BASNet\\\cite{qin2019basnet}}	&	\tabincell{c}{GateNet\\\cite{zhao2020suppress}}	&	\tabincell{c}{F$^3$Net\\\cite{wei2020f3net}}	&	\tabincell{c}{GCPANet\\\cite{chen2020global}}	&	\tabincell{c}{U$^2$Net\\\cite{qin2020u2}}	&	\tabincell{c}{SINetV2\\\cite{fan2021concealed}}	&	\tabincell{c}{PFNet\\\cite{Mei_2021_CVPR}}	&	\tabincell{c}{PSPNet\\\cite{zhao2017pyramid}}	&	\tabincell{c}{DLV3+\\\cite{chen2018encoder}}	&
        \tabincell{c}{HRNet\\\cite{WangSCJDZLMTWLX19}} &
        \tabincell{c}{BSV1\\\cite{yu2018bisenet}}	&	\tabincell{c}{ICNet\\\cite{zhao2018icnet}}	&
        \tabincell{c}{MBV3\\\cite{DBLP:conf/iccv/HowardPALSCWCTC19}} & \tabincell{c}{STDC\\\cite{Fan_2021_CVPR}}	&	\tabincell{c}{HySM\\\cite{nirkin2020hyperseg}}	&	Ours	\\
        \hline 
        \multirow{6}{*}{\begin{sideways}\tabincell{c}{\textbf{1}\\\textbf{Accessories}}\end{sideways}} & $maxF_\beta\uparrow$ &0.680 & 0.735 & 0.677 & 0.700 & 0.664 & 0.749 & 0.684 & 0.703 & 0.701 & 0.659 & 0.733 & 0.655 & 0.681 & 0.723 & 0.714 & 0.749 & 0.788 \\
        & $F^w_\beta\uparrow$ &0.576 & 0.641 & 0.572 & 0.608 & 0.560 & 0.658 & 0.606 & 0.619 & 0.614 & 0.565 & 0.652 & 0.535 & 0.590 & 0.651 & 0.631 & 0.657 & 0.716 \\
        & $~M~\downarrow$ &0.133 & 0.109 & 0.130 & 0.121 & 0.135 & 0.110 & 0.124 & 0.117 & 0.116 & 0.131 & 0.106 & 0.144 & 0.123 & 0.108 & 0.115 & 0.106 & 0.093 \\
        & $S_{\alpha}\uparrow$ &0.714 & 0.746 & 0.700 & 0.721 & 0.706 & 0.757 & 0.720 & 0.730 & 0.725 & 0.694 & 0.755 & 0.698 & 0.711 & 0.742 & 0.734 & 0.767 & 0.788 \\
        & $E_{\phi}^{m}\uparrow$ &0.761 & 0.806 & 0.770 & 0.800 & 0.759 & 0.804 & 0.794 & 0.810 & 0.800 & 0.777 & 0.818 & 0.738 & 0.786 & 0.829 & 0.814 & 0.809 & 0.837 \\
        & $HCE_\gamma\downarrow$ &547 & 549 & 571 & 612 & 682 & 562 & 679 & 634 & 662 & 580 & 581 & 688 & 585 & 684 & 630 & 547 & 432 \\
        \hline 
        \multirow{6}{*}{\begin{sideways}\tabincell{c}{\textbf{2}\\\textbf{Aircraft}}\end{sideways}} & $maxF_\beta\uparrow$ &0.823 & 0.846 & 0.788 & 0.800 & 0.781 & 0.847 & 0.804 & 0.811 & 0.816 & 0.788 & 0.831 & 0.798 & 0.814 & 0.825 & 0.791 & 0.835 & 0.886 \\
        & $F^w_\beta\uparrow$ &0.732 & 0.756 & 0.683 & 0.715 & 0.667 & 0.757 & 0.717 & 0.729 & 0.722 & 0.691 & 0.746 & 0.686 & 0.727 & 0.757 & 0.712 & 0.750 & 0.821 \\
        & $~M~\downarrow$ &0.068 & 0.063 & 0.079 & 0.069 & 0.075 & 0.064 & 0.064 & 0.067 & 0.065 & 0.076 & 0.062 & 0.075 & 0.068 & 0.056 & 0.070 & 0.062 & 0.047 \\
        & $S_{\alpha}\uparrow$ &0.829 & 0.828 & 0.779 & 0.803 & 0.791 & 0.830 & 0.807 & 0.810 & 0.814 & 0.791 & 0.830 & 0.810 & 0.817 & 0.827 & 0.800 & 0.835 & 0.872 \\
        & $E_{\phi}^{m}\uparrow$ &0.875 & 0.875 & 0.828 & 0.865 & 0.840 & 0.871 & 0.884 & 0.879 & 0.869 & 0.851 & 0.890 & 0.851 & 0.873 & 0.906 & 0.874 & 0.875 & 0.911 \\
        & $HCE_\gamma\downarrow$ &1185 & 1248 & 1153 & 1258 & 1222 & 1242 & 1241 & 1243 & 1229 & 1190 & 1448 & 1296 & 1159 & 1315 & 1314 & 1123 & 1066 \\
        \hline 
        \multirow{6}{*}{\begin{sideways}\tabincell{c}{\textbf{3}\\\textbf{Aquatic}}\end{sideways}} & $maxF_\beta\uparrow$ &0.612 & 0.681 & 0.613 & 0.613 & 0.581 & 0.691 & 0.571 & 0.649 & 0.615 & 0.601 & 0.700 & 0.563 & 0.617 & 0.672 & 0.604 & 0.654 & 0.715 \\
        & $F^w_\beta\uparrow$ &0.489 & 0.576 & 0.481 & 0.510 & 0.464 & 0.581 & 0.481 & 0.550 & 0.511 & 0.492 & 0.603 & 0.424 & 0.519 & 0.591 & 0.505 & 0.542 & 0.624 \\
        & $~M~\downarrow$ &0.119 & 0.093 & 0.109 & 0.107 & 0.124 & 0.090 & 0.124 & 0.099 & 0.103 & 0.113 & 0.085 & 0.119 & 0.103 & 0.085 & 0.104 & 0.106 & 0.080 \\
        & $S_{\alpha}\uparrow$ &0.692 & 0.728 & 0.665 & 0.687 & 0.670 & 0.738 & 0.676 & 0.716 & 0.692 & 0.673 & 0.748 & 0.658 & 0.695 & 0.729 & 0.681 & 0.713 & 0.759 \\
        & $E_{\phi}^{m}\uparrow$ &0.732 & 0.779 & 0.732 & 0.743 & 0.704 & 0.786 & 0.735 & 0.796 & 0.755 & 0.758 & 0.832 & 0.678 & 0.781 & 0.822 & 0.735 & 0.747 & 0.799 \\
        & $HCE_\gamma\downarrow$ &879 & 867 & 867 & 905 & 916 & 872 & 945 & 937 & 988 & 906 & 926 & 984 & 899 & 1009 & 938 & 839 & 710 \\
        \hline 
        \multirow{6}{*}{\begin{sideways}\tabincell{c}{\textbf{4}\\\textbf{Architecture}}\end{sideways}} & $maxF_\beta\uparrow$ &0.720 & 0.742 & 0.678 & 0.685 & 0.638 & 0.751 & 0.671 & 0.702 & 0.694 & 0.674 & 0.739 & 0.681 & 0.710 & 0.706 & 0.704 & 0.756 & 0.792 \\
        & $F^w_\beta\uparrow$ &0.610 & 0.649 & 0.570 & 0.595 & 0.528 & 0.657 & 0.587 & 0.612 & 0.601 & 0.576 & 0.649 & 0.563 & 0.621 & 0.633 & 0.622 & 0.661 & 0.713 \\
        & $~M~\downarrow$ &0.099 & 0.087 & 0.106 & 0.101 & 0.115 & 0.084 & 0.105 & 0.100 & 0.097 & 0.106 & 0.087 & 0.103 & 0.095 & 0.091 & 0.093 & 0.084 & 0.070 \\
        & $S_{\alpha}\uparrow$ &0.769 & 0.779 & 0.725 & 0.741 & 0.716 & 0.790 & 0.739 & 0.752 & 0.751 & 0.729 & 0.780 & 0.747 & 0.761 & 0.759 & 0.756 & 0.794 & 0.814 \\
        & $E_{\phi}^{m}\uparrow$ &0.803 & 0.828 & 0.779 & 0.806 & 0.759 & 0.828 & 0.813 & 0.824 & 0.808 & 0.803 & 0.841 & 0.781 & 0.821 & 0.842 & 0.829 & 0.835 & 0.849 \\
        & $HCE_\gamma\downarrow$ &1949 & 2180 & 2263 & 2368 & 2322 & 2217 & 2362 & 2418 & 2409 & 2331 & 2342 & 2525 & 2329 & 2413 & 2424 & 2053 & 1746 \\
        \hline 
        \multirow{6}{*}{\begin{sideways}\tabincell{c}{\textbf{5}\\\textbf{Artifact}}\end{sideways}} & $maxF_\beta\uparrow$ &0.721 & 0.736 & 0.687 & 0.678 & 0.640 & 0.767 & 0.648 & 0.696 & 0.702 & 0.664 & 0.741 & 0.658 & 0.713 & 0.717 & 0.693 & 0.750 & 0.805 \\
        & $F^w_\beta\uparrow$ &0.622 & 0.657 & 0.594 & 0.598 & 0.543 & 0.683 & 0.575 & 0.621 & 0.619 & 0.578 & 0.666 & 0.543 & 0.630 & 0.647 & 0.618 & 0.670 & 0.733 \\
        & $~M~\downarrow$ &0.125 & 0.107 & 0.125 & 0.128 & 0.147 & 0.100 & 0.141 & 0.125 & 0.117 & 0.134 & 0.107 & 0.144 & 0.118 & 0.114 & 0.122 & 0.107 & 0.080 \\
        & $S_{\alpha}\uparrow$ &0.758 & 0.770 & 0.725 & 0.727 & 0.708 & 0.794 & 0.712 & 0.744 & 0.747 & 0.713 & 0.777 & 0.718 & 0.751 & 0.757 & 0.735 & 0.784 & 0.822 \\
        & $E_{\phi}^{m}\uparrow$ &0.795 & 0.833 & 0.797 & 0.795 & 0.755 & 0.834 & 0.781 & 0.812 & 0.806 & 0.792 & 0.834 & 0.748 & 0.815 & 0.831 & 0.809 & 0.824 & 0.854 \\
        & $HCE_\gamma\downarrow$ &2126 & 2248 & 2572 & 2607 & 2508 & 2326 & 2454 & 2601 & 2647 & 2534 & 2494 & 2789 & 2517 & 2554 & 2613 & 2223 & 1821 \\
        \hline 
        \multirow{6}{*}{\begin{sideways}\tabincell{c}{\textbf{6}\\\textbf{Automobile}}\end{sideways}} & $maxF_\beta\uparrow$ &0.773 & 0.816 & 0.781 & 0.787 & 0.765 & 0.825 & 0.789 & 0.794 & 0.790 & 0.761 & 0.801 & 0.756 & 0.796 & 0.809 & 0.789 & 0.824 & 0.844 \\
        & $F^w_\beta\uparrow$ &0.683 & 0.741 & 0.687 & 0.708 & 0.676 & 0.752 & 0.715 & 0.719 & 0.718 & 0.680 & 0.734 & 0.659 & 0.717 & 0.748 & 0.715 & 0.745 & 0.785 \\
        & $~M~\downarrow$ &0.113 & 0.088 & 0.109 & 0.100 & 0.109 & 0.084 & 0.097 & 0.098 & 0.096 & 0.111 & 0.092 & 0.118 & 0.096 & 0.083 & 0.098 & 0.084 & 0.076 \\
        & $S_{\alpha}\uparrow$ &0.780 & 0.813 & 0.770 & 0.786 & 0.776 & 0.822 & 0.794 & 0.792 & 0.792 & 0.765 & 0.808 & 0.776 & 0.795 & 0.806 & 0.786 & 0.823 & 0.836 \\
        & $E_{\phi}^{m}\uparrow$ &0.824 & 0.865 & 0.832 & 0.850 & 0.829 & 0.868 & 0.858 & 0.862 & 0.857 & 0.842 & 0.861 & 0.820 & 0.860 & 0.879 & 0.859 & 0.868 & 0.881 \\
        & $HCE_\gamma\downarrow$ &860 & 896 & 955 & 994 & 1026 & 911 & 1037 & 1016 & 1043 & 959 & 967 & 1102 & 974 & 1056 & 1006 & 860 & 703 \\
        \hline 
        \multirow{6}{*}{\begin{sideways}\tabincell{c}{\textbf{7}\\\textbf{Electrical}}\end{sideways}} & $maxF_\beta\uparrow$ &0.625 & 0.716 & 0.656 & 0.653 & 0.584 & 0.731 & 0.625 & 0.638 & 0.638 & 0.610 & 0.691 & 0.593 & 0.662 & 0.658 & 0.653 & 0.700 & 0.778 \\
        & $F^w_\beta\uparrow$ &0.512 & 0.614 & 0.551 & 0.554 & 0.469 & 0.626 & 0.529 & 0.538 & 0.543 & 0.512 & 0.592 & 0.472 & 0.562 & 0.578 & 0.561 & 0.598 & 0.700 \\
        & $~M~\downarrow$ &0.091 & 0.065 & 0.074 & 0.073 & 0.089 & 0.064 & 0.081 & 0.082 & 0.076 & 0.083 & 0.069 & 0.090 & 0.074 & 0.072 & 0.074 & 0.070 & 0.053 \\
        & $S_{\alpha}\uparrow$ &0.730 & 0.771 & 0.728 & 0.732 & 0.701 & 0.782 & 0.715 & 0.722 & 0.728 & 0.709 & 0.760 & 0.706 & 0.742 & 0.737 & 0.731 & 0.769 & 0.808 \\
        & $E_{\phi}^{m}\uparrow$ &0.766 & 0.830 & 0.804 & 0.819 & 0.750 & 0.826 & 0.804 & 0.808 & 0.800 & 0.804 & 0.826 & 0.758 & 0.822 & 0.838 & 0.826 & 0.816 & 0.853 \\
        & $HCE_\gamma\downarrow$ &1104 & 1368 & 1333 & 1398 & 1335 & 1380 & 1358 & 1428 & 1409 & 1376 & 1501 & 1501 & 1336 & 1435 & 1421 & 1149 & 911 \\
        \hline 
        \multirow{6}{*}{\begin{sideways}\tabincell{c}{\textbf{8}\\\textbf{Electronics}}\end{sideways}} & $maxF_\beta\uparrow$ &0.721 & 0.740 & 0.688 & 0.718 & 0.658 & 0.769 & 0.712 & 0.714 & 0.715 & 0.665 & 0.733 & 0.682 & 0.723 & 0.723 & 0.712 & 0.760 & 0.801 \\
        & $F^w_\beta\uparrow$ &0.629 & 0.660 & 0.592 & 0.637 & 0.563 & 0.692 & 0.638 & 0.637 & 0.634 & 0.577 & 0.658 & 0.572 & 0.642 & 0.665 & 0.636 & 0.678 & 0.744 \\
        & $~M~\downarrow$ &0.094 & 0.089 & 0.106 & 0.098 & 0.112 & 0.080 & 0.091 & 0.096 & 0.092 & 0.108 & 0.087 & 0.108 & 0.089 & 0.086 & 0.092 & 0.084 & 0.063 \\
        & $S_{\alpha}\uparrow$ &0.780 & 0.780 & 0.737 & 0.766 & 0.739 & 0.808 & 0.769 & 0.766 & 0.769 & 0.730 & 0.784 & 0.752 & 0.771 & 0.783 & 0.764 & 0.805 & 0.834 \\
        & $E_{\phi}^{m}\uparrow$ &0.808 & 0.819 & 0.782 & 0.816 & 0.774 & 0.841 & 0.826 & 0.820 & 0.812 & 0.793 & 0.826 & 0.781 & 0.816 & 0.834 & 0.823 & 0.832 & 0.872 \\
        & $HCE_\gamma\downarrow$ &804 & 857 & 842 & 924 & 953 & 861 & 965 & 947 & 985 & 902 & 956 & 1019 & 868 & 995 & 958 & 781 & 622 \\
        \hline 
        \multirow{6}{*}{\begin{sideways}\tabincell{c}{\textbf{9}\\\textbf{Entertainment}}\end{sideways}} & $maxF_\beta\uparrow$ &0.747 & 0.784 & 0.718 & 0.716 & 0.654 & 0.774 & 0.704 & 0.738 & 0.722 & 0.699 & 0.768 & 0.727 & 0.746 & 0.746 & 0.730 & 0.791 & 0.831 \\
        & $F^w_\beta\uparrow$ &0.628 & 0.681 & 0.603 & 0.615 & 0.532 & 0.671 & 0.605 & 0.639 & 0.615 & 0.592 & 0.671 & 0.600 & 0.648 & 0.663 & 0.640 & 0.688 & 0.748 \\
        & $~M~\downarrow$ &0.110 & 0.093 & 0.111 & 0.111 & 0.126 & 0.095 & 0.110 & 0.105 & 0.106 & 0.117 & 0.094 & 0.112 & 0.100 & 0.097 & 0.103 & 0.093 & 0.071 \\
        & $S_{\alpha}\uparrow$ &0.768 & 0.786 & 0.737 & 0.743 & 0.713 & 0.783 & 0.742 & 0.761 & 0.745 & 0.729 & 0.781 & 0.759 & 0.769 & 0.767 & 0.754 & 0.799 & 0.827 \\
        & $E_{\phi}^{m}\uparrow$ &0.802 & 0.839 & 0.798 & 0.821 & 0.760 & 0.834 & 0.830 & 0.837 & 0.816 & 0.814 & 0.850 & 0.801 & 0.836 & 0.852 & 0.840 & 0.838 & 0.872 \\
        & $HCE_\gamma\downarrow$ &1644 & 1793 & 1837 & 1862 & 1834 & 1872 & 1849 & 1904 & 1907 & 1838 & 1969 & 2029 & 1819 & 1920 & 1870 & 1643 & 1369 \\
        \hline 
        \multirow{6}{*}{\begin{sideways}\tabincell{c}{\textbf{10}\\\textbf{Frame}}\end{sideways}} & $maxF_\beta\uparrow$ &0.681 & 0.718 & 0.678 & 0.651 & 0.596 & 0.742 & 0.629 & 0.671 & 0.680 & 0.638 & 0.687 & 0.643 & 0.675 & 0.696 & 0.695 & 0.724 & 0.783 \\
        & $F^w_\beta\uparrow$ &0.564 & 0.625 & 0.573 & 0.561 & 0.482 & 0.639 & 0.543 & 0.576 & 0.586 & 0.547 & 0.597 & 0.513 & 0.581 & 0.621 & 0.610 & 0.619 & 0.702 \\
        & $~M~\downarrow$ &0.097 & 0.080 & 0.086 & 0.093 & 0.113 & 0.075 & 0.104 & 0.093 & 0.088 & 0.097 & 0.087 & 0.104 & 0.087 & 0.082 & 0.083 & 0.082 & 0.064 \\
        & $S_{\alpha}\uparrow$ &0.757 & 0.787 & 0.750 & 0.745 & 0.717 & 0.800 & 0.732 & 0.754 & 0.758 & 0.735 & 0.767 & 0.735 & 0.758 & 0.761 & 0.759 & 0.786 & 0.826 \\
        & $E_{\phi}^{m}\uparrow$ &0.791 & 0.832 & 0.818 & 0.810 & 0.752 & 0.843 & 0.796 & 0.819 & 0.821 & 0.812 & 0.819 & 0.777 & 0.827 & 0.845 & 0.842 & 0.824 & 0.863 \\
        & $HCE_\gamma\downarrow$ &1066 & 1169 & 1248 & 1317 & 1311 & 1187 & 1318 & 1354 & 1380 & 1266 & 1294 & 1425 & 1258 & 1371 & 1318 & 1122 & 850 \\
        \hline 
        \multirow{6}{*}{\begin{sideways}\tabincell{c}{\textbf{11}\\\textbf{Furniture}}\end{sideways}} & $maxF_\beta\uparrow$ &0.655 & 0.721 & 0.662 & 0.670 & 0.629 & 0.725 & 0.664 & 0.680 & 0.675 & 0.644 & 0.706 & 0.623 & 0.670 & 0.702 & 0.680 & 0.718 & 0.773 \\
        & $F^w_\beta\uparrow$ &0.549 & 0.636 & 0.558 & 0.580 & 0.525 & 0.636 & 0.583 & 0.593 & 0.586 & 0.553 & 0.622 & 0.506 & 0.583 & 0.629 & 0.597 & 0.626 & 0.695 \\
        & $~M~\downarrow$ &0.119 & 0.090 & 0.109 & 0.106 & 0.121 & 0.089 & 0.109 & 0.103 & 0.103 & 0.111 & 0.095 & 0.126 & 0.103 & 0.090 & 0.102 & 0.095 & 0.076 \\
        & $S_{\alpha}\uparrow$ &0.725 & 0.768 & 0.715 & 0.730 & 0.711 & 0.773 & 0.733 & 0.741 & 0.736 & 0.710 & 0.761 & 0.705 & 0.734 & 0.754 & 0.736 & 0.768 & 0.804 \\
        & $E_{\phi}^{m}\uparrow$ &0.764 & 0.822 & 0.787 & 0.796 & 0.761 & 0.819 & 0.803 & 0.805 & 0.799 & 0.794 & 0.813 & 0.750 & 0.803 & 0.834 & 0.811 & 0.811 & 0.842 \\
        & $HCE_\gamma\downarrow$ &871 & 904 & 951 & 1001 & 1012 & 914 & 1012 & 1035 & 1044 & 959 & 1018 & 1120 & 978 & 1048 & 1020 & 862 & 671 \\
        \hline
	\end{tabular}
\end{table*}

\begin{table*}[thbp]
\scriptsize
\centering
	\caption{\small PART-II: Quantitative evaluation on our validation, DIS-VD, and test sets, DIS-TE(1-4), based on groups. ResNet18=R-18. ResNet34=R-34. ResNet50=R-50. Res2Net50=R2-50. DeepLab-V3+=DLV3+. BiseNetV1=BSV1. STDC813=S-813. EffiNetB1=E-B1. MobileNetV3-Large=MBV3. HyperSeg-M=HySM.
	}\label{tab:compSOA-2}
    \renewcommand{\arraystretch}{0.99}
    \setlength\tabcolsep{1.7pt}
	\begin{tabular}{c|r|cccccc|cc|ccc|ccccc|c}
		\hline
        Dataset	&	Metric	&	\tabincell{c}{UNet\\\cite{ronneberger2015u}}	&	\tabincell{c}{BASNet\\\cite{qin2019basnet}}	&	\tabincell{c}{GateNet\\\cite{zhao2020suppress}}	&	\tabincell{c}{F$^3$Net\\\cite{wei2020f3net}}	&	\tabincell{c}{GCPANet\\\cite{chen2020global}}	&	\tabincell{c}{U$^2$Net\\\cite{qin2020u2}}	&	\tabincell{c}{SINetV2\\\cite{fan2021concealed}}	&	\tabincell{c}{PFNet\\\cite{Mei_2021_CVPR}}	&	\tabincell{c}{PSPNet\\\cite{zhao2017pyramid}}	&	\tabincell{c}{DLV3+\\\cite{chen2018encoder}}	&
        \tabincell{c}{HRNet\\\cite{WangSCJDZLMTWLX19}} &
        \tabincell{c}{BSV1\\\cite{yu2018bisenet}}	&	\tabincell{c}{ICNet\\\cite{zhao2018icnet}}	&
        \tabincell{c}{MBV3\\\cite{DBLP:conf/iccv/HowardPALSCWCTC19}} & \tabincell{c}{STDC\\\cite{Fan_2021_CVPR}}	&	\tabincell{c}{HySM\\\cite{nirkin2020hyperseg}}	&	Ours	\\
        \hline 
        \multirow{6}{*}{\begin{sideways}\tabincell{c}{\textbf{12}\\\textbf{Graphics}}\end{sideways}} & $maxF_\beta\uparrow$ &0.750 & 0.719 & 0.685 & 0.663 & 0.524 & 0.746 & 0.568 & 0.645 & 0.646 & 0.621 & 0.671 & 0.575 & 0.681 & 0.616 & 0.647 & 0.732 & 0.780 \\
        & $F^w_\beta\uparrow$ &0.654 & 0.628 & 0.598 & 0.584 & 0.431 & 0.653 & 0.496 & 0.569 & 0.566 & 0.540 & 0.585 & 0.473 & 0.606 & 0.566 & 0.578 & 0.647 & 0.706 \\
        & $~M~\downarrow$ &0.061 & 0.064 & 0.066 & 0.069 & 0.094 & 0.057 & 0.088 & 0.078 & 0.067 & 0.073 & 0.074 & 0.096 & 0.064 & 0.065 & 0.068 & 0.059 & 0.049 \\
        & $S_{\alpha}\uparrow$ &0.825 & 0.800 & 0.784 & 0.772 & 0.703 & 0.823 & 0.717 & 0.754 & 0.772 & 0.752 & 0.772 & 0.719 & 0.790 & 0.750 & 0.763 & 0.814 & 0.839 \\
        & $E_{\phi}^{m}\uparrow$ &0.835 & 0.831 & 0.835 & 0.843 & 0.726 & 0.834 & 0.795 & 0.827 & 0.798 & 0.817 & 0.819 & 0.740 & 0.828 & 0.847 & 0.865 & 0.836 & 0.873 \\
        & $HCE_\gamma\downarrow$ &670 & 976 & 1009 & 1268 & 1403 & 938 & 1423 & 1294 & 1447 & 1201 & 990 & 1425 & 1122 & 1457 & 1331 & 824 & 621 \\
        \hline 
        \multirow{6}{*}{\begin{sideways}\tabincell{c}{\textbf{13}\\\textbf{Insect}}\end{sideways}} & $maxF_\beta\uparrow$ &0.673 & 0.681 & 0.641 & 0.627 & 0.554 & 0.718 & 0.608 & 0.634 & 0.637 & 0.617 & 0.706 & 0.620 & 0.650 & 0.700 & 0.643 & 0.692 & 0.762 \\
        & $F^w_\beta\uparrow$ &0.552 & 0.586 & 0.530 & 0.537 & 0.442 & 0.617 & 0.523 & 0.541 & 0.541 & 0.522 & 0.617 & 0.482 & 0.552 & 0.629 & 0.557 & 0.592 & 0.683 \\
        & $~M~\downarrow$ &0.073 & 0.065 & 0.071 & 0.070 & 0.089 & 0.058 & 0.076 & 0.075 & 0.069 & 0.074 & 0.061 & 0.075 & 0.068 & 0.058 & 0.069 & 0.062 & 0.049 \\
        & $S_{\alpha}\uparrow$ &0.766 & 0.766 & 0.733 & 0.738 & 0.694 & 0.786 & 0.724 & 0.737 & 0.740 & 0.728 & 0.785 & 0.725 & 0.747 & 0.776 & 0.743 & 0.783 & 0.820 \\
        & $E_{\phi}^{m}\uparrow$ &0.804 & 0.821 & 0.781 & 0.804 & 0.753 & 0.827 & 0.810 & 0.803 & 0.817 & 0.817 & 0.844 & 0.748 & 0.825 & 0.863 & 0.820 & 0.809 & 0.860 \\
        & $HCE_\gamma\downarrow$ &570 & 595 & 604 & 656 & 683 & 592 & 701 & 663 & 714 & 636 & 622 & 700 & 609 & 713 & 667 & 574 & 488 \\
        \hline 
        \multirow{6}{*}{\begin{sideways}\tabincell{c}{\textbf{14}\\\textbf{Kitchenware}}\end{sideways}} & $maxF_\beta\uparrow$ &0.704 & 0.754 & 0.678 & 0.697 & 0.688 & 0.734 & 0.713 & 0.708 & 0.692 & 0.661 & 0.739 & 0.667 & 0.689 & 0.730 & 0.702 & 0.749 & 0.771 \\
        & $F^w_\beta\uparrow$ &0.588 & 0.654 & 0.555 & 0.596 & 0.578 & 0.633 & 0.620 & 0.608 & 0.587 & 0.550 & 0.649 & 0.545 & 0.587 & 0.647 & 0.606 & 0.657 & 0.685 \\
        & $~M~\downarrow$ &0.167 & 0.144 & 0.174 & 0.163 & 0.170 & 0.151 & 0.152 & 0.160 & 0.167 & 0.178 & 0.143 & 0.178 & 0.166 & 0.144 & 0.159 & 0.140 & 0.128 \\
        & $S_{\alpha}\uparrow$ &0.704 & 0.733 & 0.662 & 0.690 & 0.691 & 0.723 & 0.712 & 0.698 & 0.680 & 0.653 & 0.729 & 0.679 & 0.688 & 0.729 & 0.697 & 0.743 & 0.763 \\
        & $E_{\phi}^{m}\uparrow$ &0.737 & 0.777 & 0.721 & 0.754 & 0.742 & 0.761 & 0.777 & 0.764 & 0.736 & 0.731 & 0.798 & 0.725 & 0.753 & 0.795 & 0.764 & 0.786 & 0.798 \\
        & $HCE_\gamma\downarrow$ &541 & 536 & 554 & 574 & 579 & 536 & 602 & 583 & 588 & 543 & 608 & 637 & 540 & 608 & 571 & 484 & 367 \\
        \hline 
        \multirow{6}{*}{\begin{sideways}\tabincell{c}{\textbf{15}\\\textbf{Machine}}\end{sideways}} & $maxF_\beta\uparrow$ &0.798 & 0.807 & 0.744 & 0.777 & 0.746 & 0.845 & 0.778 & 0.767 & 0.800 & 0.766 & 0.842 & 0.755 & 0.812 & 0.812 & 0.782 & 0.818 & 0.869 \\
        & $F^w_\beta\uparrow$ &0.692 & 0.713 & 0.629 & 0.676 & 0.638 & 0.755 & 0.695 & 0.676 & 0.710 & 0.666 & 0.760 & 0.639 & 0.722 & 0.738 & 0.694 & 0.727 & 0.801 \\
        & $~M~\downarrow$ &0.126 & 0.119 & 0.147 & 0.131 & 0.145 & 0.100 & 0.124 & 0.131 & 0.118 & 0.138 & 0.100 & 0.147 & 0.116 & 0.111 & 0.123 & 0.116 & 0.089 \\
        & $S_{\alpha}\uparrow$ &0.764 & 0.771 & 0.701 & 0.739 & 0.728 & 0.809 & 0.761 & 0.736 & 0.770 & 0.729 & 0.802 & 0.739 & 0.773 & 0.780 & 0.747 & 0.783 & 0.842 \\
        & $E_{\phi}^{m}\uparrow$ &0.812 & 0.833 & 0.781 & 0.816 & 0.786 & 0.851 & 0.844 & 0.824 & 0.843 & 0.821 & 0.870 & 0.779 & 0.848 & 0.857 & 0.835 & 0.835 & 0.881 \\
        & $HCE_\gamma\downarrow$ &1544 & 1687 & 1728 & 1846 & 1849 & 1693 & 1910 & 1860 & 1925 & 1787 & 1937 & 1987 & 1799 & 1957 & 1899 & 1589 & 1322 \\
        \hline 
        \multirow{6}{*}{\begin{sideways}\tabincell{c}{\textbf{16}\\\textbf{Music}\\\textbf{Instrument}}\end{sideways}} & $maxF_\beta\uparrow$ &0.748 & 0.809 & 0.740 & 0.777 & 0.756 & 0.817 & 0.775 & 0.777 & 0.777 & 0.752 & 0.808 & 0.748 & 0.774 & 0.811 & 0.777 & 0.829 & 0.852 \\
        & $F^w_\beta\uparrow$ &0.643 & 0.726 & 0.636 & 0.691 & 0.660 & 0.734 & 0.699 & 0.698 & 0.690 & 0.656 & 0.730 & 0.640 & 0.689 & 0.739 & 0.698 & 0.745 & 0.783 \\
        & $~M~\downarrow$ &0.159 & 0.123 & 0.163 & 0.137 & 0.145 & 0.115 & 0.127 & 0.133 & 0.139 & 0.154 & 0.117 & 0.156 & 0.140 & 0.113 & 0.135 & 0.114 & 0.101 \\
        & $S_{\alpha}\uparrow$ &0.732 & 0.781 & 0.706 & 0.753 & 0.749 & 0.790 & 0.767 & 0.761 & 0.749 & 0.722 & 0.787 & 0.736 & 0.750 & 0.782 & 0.755 & 0.799 & 0.820 \\
        & $E_{\phi}^{m}\uparrow$ &0.775 & 0.825 & 0.764 & 0.811 & 0.792 & 0.834 & 0.826 & 0.818 & 0.809 & 0.796 & 0.842 & 0.771 & 0.809 & 0.848 & 0.814 & 0.828 & 0.853 \\
        & $HCE_\gamma\downarrow$ &671 & 683 & 653 & 693 & 708 & 705 & 735 & 713 & 732 & 678 & 791 & 796 & 687 & 771 & 748 & 598 & 492 \\
        \hline 
        \multirow{6}{*}{\begin{sideways}\tabincell{c}{\textbf{17}\\\textbf{ Non-motor}\\\textbf{Vehicle}}\end{sideways}} & $maxF_\beta\uparrow$ &0.762 & 0.800 & 0.755 & 0.761 & 0.718 & 0.803 & 0.740 & 0.755 & 0.774 & 0.748 & 0.791 & 0.731 & 0.764 & 0.779 & 0.768 & 0.794 & 0.840 \\
        & $F^w_\beta\uparrow$ &0.662 & 0.719 & 0.658 & 0.674 & 0.612 & 0.722 & 0.654 & 0.673 & 0.687 & 0.660 & 0.713 & 0.620 & 0.683 & 0.709 & 0.691 & 0.710 & 0.774 \\
        & $~M~\downarrow$ &0.100 & 0.086 & 0.103 & 0.100 & 0.118 & 0.086 & 0.107 & 0.101 & 0.095 & 0.101 & 0.086 & 0.113 & 0.095 & 0.087 & 0.093 & 0.088 & 0.068 \\
        & $S_{\alpha}\uparrow$ &0.788 & 0.816 & 0.767 & 0.784 & 0.759 & 0.817 & 0.770 & 0.781 & 0.791 & 0.769 & 0.812 & 0.768 & 0.790 & 0.800 & 0.787 & 0.815 & 0.846 \\
        & $E_{\phi}^{m}\uparrow$ &0.839 & 0.870 & 0.836 & 0.853 & 0.807 & 0.866 & 0.845 & 0.852 & 0.857 & 0.852 & 0.870 & 0.811 & 0.857 & 0.874 & 0.863 & 0.859 & 0.891 \\
        & $HCE_\gamma\downarrow$ &1956 & 2098 & 2134 & 2219 & 2217 & 2121 & 2269 & 2293 & 2274 & 2169 & 2314 & 2319 & 2161 & 2334 & 2245 & 1971 & 1623 \\
        \hline 
        \multirow{6}{*}{\begin{sideways}\tabincell{c}{\textbf{18}\\\textbf{Plant}}\end{sideways}} & $maxF_\beta\uparrow$ &0.685 & 0.745 & 0.690 & 0.685 & 0.680 & 0.771 & 0.696 & 0.701 & 0.723 & 0.703 & 0.755 & 0.642 & 0.718 & 0.743 & 0.706 & 0.785 & 0.766 \\
        & $F^w_\beta\uparrow$ &0.566 & 0.637 & 0.569 & 0.576 & 0.564 & 0.665 & 0.589 & 0.602 & 0.623 & 0.595 & 0.658 & 0.500 & 0.621 & 0.654 & 0.597 & 0.689 & 0.665 \\
        & $~M~\downarrow$ &0.144 & 0.119 & 0.138 & 0.138 & 0.145 & 0.111 & 0.141 & 0.134 & 0.126 & 0.131 & 0.111 & 0.153 & 0.125 & 0.112 & 0.136 & 0.104 & 0.109 \\
        & $S_{\alpha}\uparrow$ &0.697 & 0.730 & 0.689 & 0.695 & 0.685 & 0.761 & 0.703 & 0.696 & 0.727 & 0.700 & 0.752 & 0.662 & 0.720 & 0.737 & 0.693 & 0.779 & 0.764 \\
        & $E_{\phi}^{m}\uparrow$ &0.749 & 0.778 & 0.749 & 0.755 & 0.748 & 0.787 & 0.758 & 0.774 & 0.790 & 0.783 & 0.810 & 0.707 & 0.801 & 0.804 & 0.762 & 0.804 & 0.779 \\
        & $HCE_\gamma\downarrow$ &9194 & 9174 & 10036 & 10164 & 10488 & 9062 & 10268 & 10137 & 10231 & 9910 & 9615 & 10444 & 9798 & 10309 & 10230 & 8334 & 8563 \\
        \hline 
        \multirow{6}{*}{\begin{sideways}\tabincell{c}{\textbf{19}\\\textbf{Ship}}\end{sideways}} & $maxF_\beta\uparrow$ &0.773 & 0.793 & 0.739 & 0.747 & 0.726 & 0.792 & 0.730 & 0.760 & 0.769 & 0.756 & 0.779 & 0.761 & 0.772 & 0.785 & 0.744 & 0.791 & 0.834 \\
        & $F^w_\beta\uparrow$ &0.686 & 0.705 & 0.632 & 0.660 & 0.614 & 0.713 & 0.648 & 0.672 & 0.676 & 0.657 & 0.698 & 0.653 & 0.690 & 0.711 & 0.659 & 0.711 & 0.766 \\
        & $~M~\downarrow$ &0.095 & 0.095 & 0.114 & 0.107 & 0.116 & 0.089 & 0.108 & 0.103 & 0.103 & 0.107 & 0.098 & 0.104 & 0.098 & 0.085 & 0.108 & 0.091 & 0.069 \\
        & $S_{\alpha}\uparrow$ &0.796 & 0.796 & 0.742 & 0.760 & 0.741 & 0.804 & 0.753 & 0.770 & 0.775 & 0.758 & 0.784 & 0.772 & 0.787 & 0.790 & 0.757 & 0.806 & 0.840 \\
        & $E_{\phi}^{m}\uparrow$ &0.840 & 0.842 & 0.793 & 0.823 & 0.785 & 0.849 & 0.838 & 0.837 & 0.828 & 0.826 & 0.846 & 0.811 & 0.846 & 0.870 & 0.831 & 0.848 & 0.880 \\
        & $HCE_\gamma\downarrow$ &3193 & 3341 & 3233 & 3242 & 3225 & 3355 & 3183 & 3265 & 3189 & 3178 & 3443 & 3454 & 3134 & 3381 & 3334 & 3046 & 2951 \\
        \hline 
        \multirow{6}{*}{\begin{sideways}\tabincell{c}{\textbf{20}\\\textbf{Sports}}\end{sideways}} & $maxF_\beta\uparrow$ &0.699 & 0.721 & 0.674 & 0.675 & 0.637 & 0.745 & 0.661 & 0.687 & 0.685 & 0.639 & 0.724 & 0.679 & 0.676 & 0.727 & 0.684 & 0.744 & 0.788 \\
        & $F^w_\beta\uparrow$ &0.596 & 0.629 & 0.572 & 0.583 & 0.526 & 0.651 & 0.573 & 0.590 & 0.594 & 0.547 & 0.637 & 0.554 & 0.583 & 0.654 & 0.597 & 0.647 & 0.714 \\
        & $~M~\downarrow$ &0.076 & 0.065 & 0.074 & 0.074 & 0.081 & 0.059 & 0.077 & 0.075 & 0.072 & 0.081 & 0.064 & 0.078 & 0.072 & 0.059 & 0.072 & 0.062 & 0.051 \\
        & $S_{\alpha}\uparrow$ &0.766 & 0.778 & 0.743 & 0.747 & 0.728 & 0.797 & 0.740 & 0.748 & 0.748 & 0.724 & 0.784 & 0.751 & 0.752 & 0.780 & 0.750 & 0.795 & 0.827 \\
        & $E_{\phi}^{m}\uparrow$ &0.807 & 0.822 & 0.805 & 0.825 & 0.777 & 0.832 & 0.821 & 0.825 & 0.820 & 0.803 & 0.831 & 0.801 & 0.816 & 0.868 & 0.836 & 0.838 & 0.860 \\
        & $HCE_\gamma\downarrow$ &1137 & 1283 & 1274 & 1329 & 1247 & 1315 & 1274 & 1355 & 1323 & 1297 & 1450 & 1401 & 1306 & 1352 & 1343 & 1180 & 934 \\
        \hline 
        \multirow{6}{*}{\begin{sideways}\tabincell{c}{\textbf{21}\\\textbf{Tool}}\end{sideways}} & $maxF_\beta\uparrow$ &0.656 & 0.714 & 0.649 & 0.678 & 0.643 & 0.719 & 0.670 & 0.683 & 0.679 & 0.628 & 0.700 & 0.628 & 0.670 & 0.697 & 0.680 & 0.717 & 0.757 \\
        & $F^w_\beta\uparrow$ &0.538 & 0.622 & 0.543 & 0.582 & 0.533 & 0.624 & 0.581 & 0.589 & 0.588 & 0.532 & 0.612 & 0.505 & 0.573 & 0.623 & 0.592 & 0.611 & 0.676 \\
        & $~M~\downarrow$ &0.100 & 0.082 & 0.095 & 0.089 & 0.100 & 0.080 & 0.094 & 0.090 & 0.086 & 0.101 & 0.086 & 0.104 & 0.091 & 0.081 & 0.087 & 0.082 & 0.071 \\
        & $S_{\alpha}\uparrow$ &0.733 & 0.771 & 0.721 & 0.743 & 0.727 & 0.773 & 0.739 & 0.746 & 0.752 & 0.708 & 0.759 & 0.719 & 0.740 & 0.758 & 0.739 & 0.771 & 0.797 \\
        & $E_{\phi}^{m}\uparrow$ &0.784 & 0.829 & 0.797 & 0.822 & 0.785 & 0.831 & 0.815 & 0.822 & 0.820 & 0.802 & 0.827 & 0.769 & 0.815 & 0.842 & 0.836 & 0.823 & 0.844 \\
        & $HCE_\gamma\downarrow$ &568 & 589 & 620 & 659 & 673 & 602 & 689 & 673 & 689 & 632 & 662 & 724 & 625 & 707 & 660 & 554 & 433 \\
        \hline 
        \multirow{6}{*}{\begin{sideways}\tabincell{c}{\textbf{22}\\\textbf{Weapon}}\end{sideways}} & $maxF_\beta\uparrow$ &0.763 & 0.805 & 0.728 & 0.787 & 0.765 & 0.816 & 0.780 & 0.799 & 0.798 & 0.747 & 0.812 & 0.757 & 0.773 & 0.794 & 0.785 & 0.806 & 0.848 \\
        & $F^w_\beta\uparrow$ &0.672 & 0.726 & 0.616 & 0.706 & 0.668 & 0.737 & 0.706 & 0.717 & 0.718 & 0.654 & 0.743 & 0.657 & 0.689 & 0.728 & 0.707 & 0.730 & 0.794 \\
        & $~M~\downarrow$ &0.108 & 0.090 & 0.124 & 0.097 & 0.106 & 0.087 & 0.096 & 0.093 & 0.091 & 0.113 & 0.084 & 0.108 & 0.099 & 0.088 & 0.096 & 0.085 & 0.071 \\
        & $S_{\alpha}\uparrow$ &0.784 & 0.802 & 0.718 & 0.788 & 0.775 & 0.814 & 0.790 & 0.797 & 0.794 & 0.750 & 0.816 & 0.778 & 0.776 & 0.802 & 0.784 & 0.820 & 0.850 \\
        & $E_{\phi}^{m}\uparrow$ &0.822 & 0.861 & 0.790 & 0.843 & 0.833 & 0.855 & 0.857 & 0.861 & 0.856 & 0.828 & 0.870 & 0.826 & 0.838 & 0.864 & 0.854 & 0.867 & 0.899 \\
        & $HCE_\gamma\downarrow$ &793 & 826 & 849 & 888 & 899 & 845 & 923 & 902 & 928 & 864 & 914 & 969 & 861 & 937 & 899 & 779 & 649 \\
        \hline 
        \multirow{6}{*}{\begin{sideways}\tabincell{c}{\textbf{All}\\\textbf{VD+TE(1-4)}}\end{sideways}} & $maxF_\beta\uparrow$ &0.705 & 0.748 & 0.691 & 0.700 & 0.658 & 0.758 & 0.687 & 0.708 & 0.706 & 0.675 & 0.739 & 0.671 & 0.709 & 0.726 & 0.708 & 0.752 & 0.798 \\
        & $F^w_\beta\uparrow$ &0.600 & 0.659 & 0.587 & 0.611 & 0.551 & 0.668 & 0.604 & 0.620 & 0.617 & 0.581 & 0.654 & 0.556 & 0.620 & 0.655 & 0.625 & 0.660 & 0.724 \\
        & $~M~\downarrow$ &0.105 & 0.087 & 0.104 & 0.099 & 0.113 & 0.085 & 0.102 & 0.099 & 0.096 & 0.107 & 0.088 & 0.112 & 0.096 & 0.087 & 0.095 & 0.087 & 0.071 \\
        & $S_{\alpha}\uparrow$ &0.756 & 0.780 & 0.730 & 0.747 & 0.726 & 0.789 & 0.743 & 0.753 & 0.753 & 0.727 & 0.778 & 0.735 & 0.756 & 0.768 & 0.751 & 0.788 & 0.818 \\
        & $E_{\phi}^{m}\uparrow$ &0.796 & 0.832 & 0.794 & 0.815 & 0.774 & 0.833 & 0.817 & 0.824 & 0.816 & 0.807 & 0.837 & 0.776 & 0.822 & 0.849 & 0.829 & 0.830 & 0.857 \\
        & $HCE_\gamma\downarrow$ &1228 & 1330 & 1368 & 1433 & 1425 & 1348 & 1441 & 1461 & 1470 & 1394 & 1457 & 1541 & 1387 & 1489 & 1459 & 1239 & 1035 \\
        \hline
	\end{tabular}
\end{table*}

\noindent 
\textbf{Performance comparisons among different test sets.} 
performance analysis based on the targets' complexities for demonstrating the importance of our newly proposed $HCE_\gamma\downarrow$ metric. 
As shown in Table~\ref{tab:compSOA}, our model achieves different performances on the four testing sets, obtained by ordering (ascending) and splitting the whole test set according to the structural complexities of the to-be-segmented objects. 
However, except for our newly proposed $HCE_\gamma\downarrow$, other metrics, such as $maxF_\beta\uparrow$, $F^w_\beta\uparrow$, $~M\downarrow$, $S_{\alpha}\uparrow$ and $E_{\phi}^{m}\uparrow$, of DIS-TE1, DIS-TE2, DIS-TE3, and DIS-TE4 show no strong (negative or positive) correlations with respect to the shape complexities. 
For example, $M$ of our model on these DIS-TE1 (0.074) and DIS-TE4 (0.072) are very close. The $maxF_\beta\uparrow$, $F^w_\beta\uparrow$, $S_{\alpha}\uparrow$ and $E_{\phi}^{m}\uparrow$ of DIS-TE4 are even greater than those of DIS-TE1, which probably provides misleading information that DIS-TE4 is less challenging than DIS-TE1. 
On the contrary, the $HCE_\gamma\downarrow$ of our model on DIS-TE1 and DIS-TE4 are 149 and 2,888, respectively. 
That indicates the cost for correcting the predictions of DIS-TE4 is around 20 times more than that of correcting predictions on DIS-TE1, which is consistent with the complexities illustrated in Table~\ref{tab:te14}. 
It means our $HCE_\gamma\downarrow$ can correctly describe the correlations between prediction quality and the shape complexities. 
Thus, it can assess the human interventions needed when applying the models to real-world applications. 
We can get similar observations from the evaluation scores of other models on different test sets, which further proves the importance of our $HCE_\gamma\downarrow$ in evaluating highly accurate dichotomous image segmentation results. 
It is worth noting that the weak correlations between the conventional metrics and the shape complexities of different test sets are partial because image context complexity also plays a vital role in determining the segmentation difficulties. 
But this factor is hard to be quantified and has relatively less impact on the labeling workloads. 
Therefore, it is not considered in this work and will be studied in the future. 
In addition, performance comparisons of different models based on different groups are illustrated in Table~\ref{tab:compSOA-1} and \ref{tab:compSOA-2}, from which the per-group segmentation difficulties and performance can be found.   

\begin{figure*}[thbp]
    \centering
    \includegraphics[width=1\textwidth]{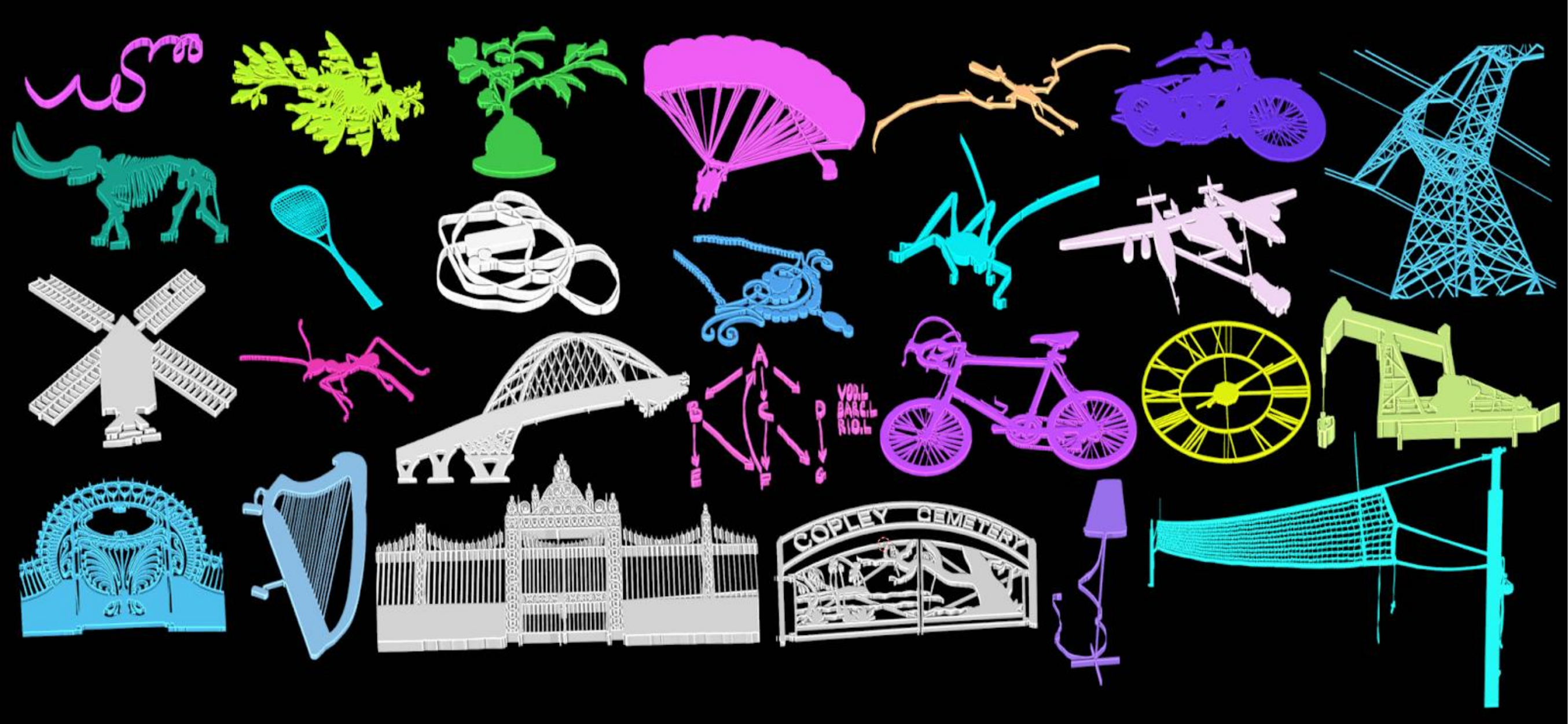}
    \caption{\small 3D models built upon the ground truth masks sampled from DIS5K by the ``Extrude'' operation in Blender.}
    \label{fig:DIS-3d}
\end{figure*}
\begin{figure*}[thbp]
    \centering
    \includegraphics[width=0.96\linewidth]{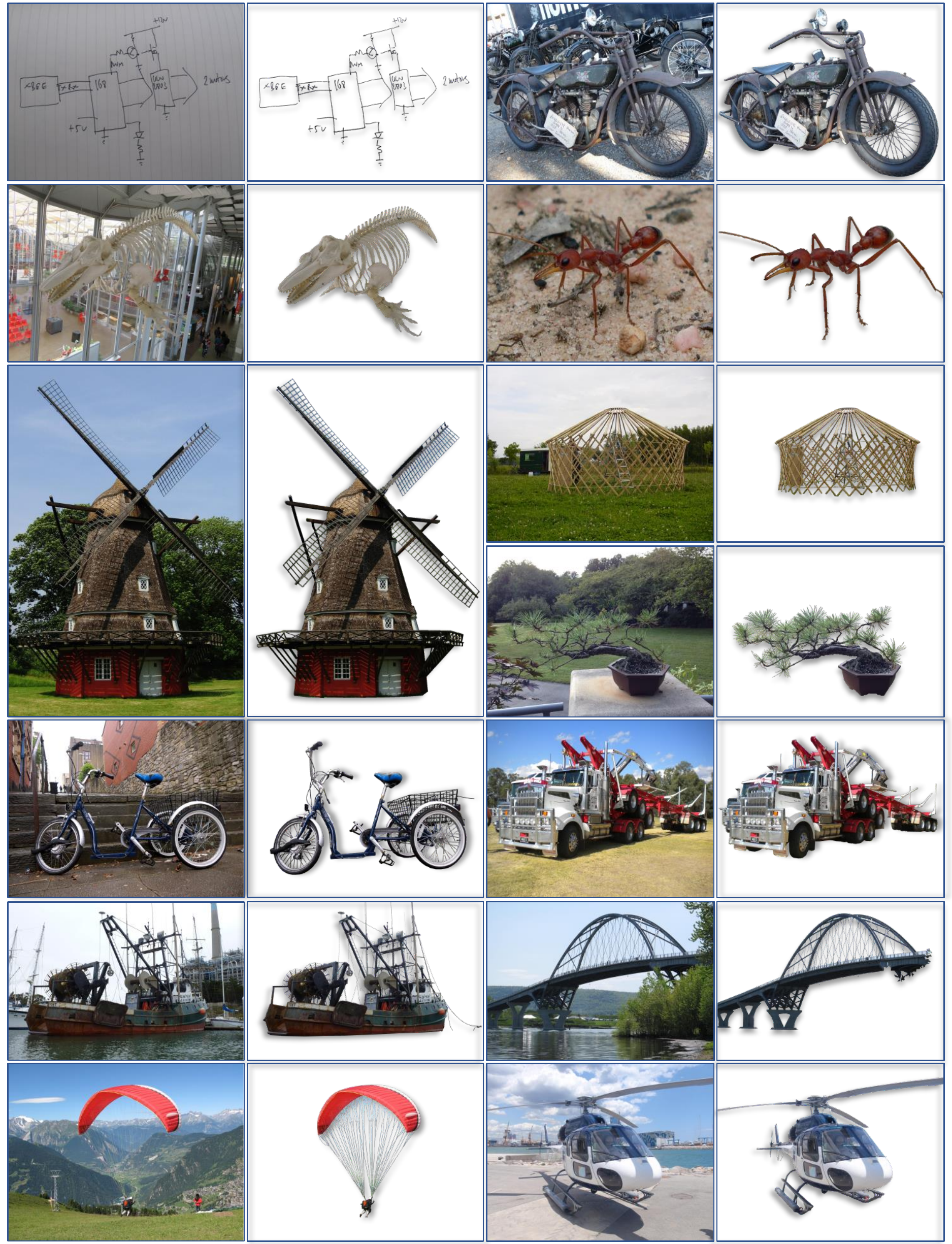}
    \caption{\small Comparisons between the original images and their backgrounds-removed correspondences generated from our DIS5K.}
    \label{fig:tm_rmbg}
\end{figure*}

\noindent 
\textbf{Effectiveness of Our Intermediate Supervision}
To further demonstrate the effectiveness of our intermediate supervision, we show the training loss and validation mean absolute error $~M\downarrow$ curves of our adapted U$^2$-Net with and without our intermediate supervisions in \figref{fig:tm_curves}. The top part of \figref{fig:tm_curves} shows the training loss of the last side output, which is taken as the final result in the inference stage. As can be seen, the models with intermediate supervisions converge faster before around 10,000 iterations. Later, the model without intermediate supervisions gradually produces a lower loss. These curves demonstrate that our intermediate supervision plays a typical role of regularizer for reducing the probability of over-fitting. The bottom plot of \figref{fig:tm_curves} shows that our intermediate supervision significantly decreases the $~M\downarrow$ on the validation set, which validates its effectiveness in performance improvement. 

\section{Applications}

Our DIS task will benefit both academia and industrious. 
In addition to the DIS task, we believe that our highly accurate large-scale DIS5K dataset can also be used in various related research fields, such as: 
\begin{itemize}
    \item providing pre-trained segmentation models for other specific object segmentation tasks as well as facilitating the downstream tasks, such as image matting, editing, and so on; 
    \item the subsets of DIS5K can be used for fast prototyping of different segmentation tasks; 
    \item providing materials and examples for shape and structure analysis in graphics and topology; 
    \item high resolution fine-grained image classification; 
    \item segmentation guided super-resolution and image processing; 
    \item synthesizing more composite images with diversified backgrounds for more robust image segmentation; 
    \item edge, boundary or contour detection, \etc.
\end{itemize}
Thanks to the high resolution and accurate labeling, many samples in our DIS5K show high artistic and aesthetic values. Fig.~\ref{fig:intro} shows the comparison between the original ship image with cluttered background and the background-removed image with perspective transforms (See more samples in Fig.~\ref{fig:tm_rmbg}). As can be seen, compared with the original image, the background-removed image shows higher aesthetic values and good usability, which can even be directly used as:  
\begin{itemize}
    \item materials of art design, image and video editing;
    \item backgrounds of posters or slides, wall papers of cellphones, tablets, desktops; 
    \item materials for 3D modeling, as shown in Fig.~\ref{fig:DIS-3d} (A demo video is also attached).
\end{itemize}


\begin{figure*}[t!]
    \centering
    \begin{overpic}[width= 0.9\linewidth]{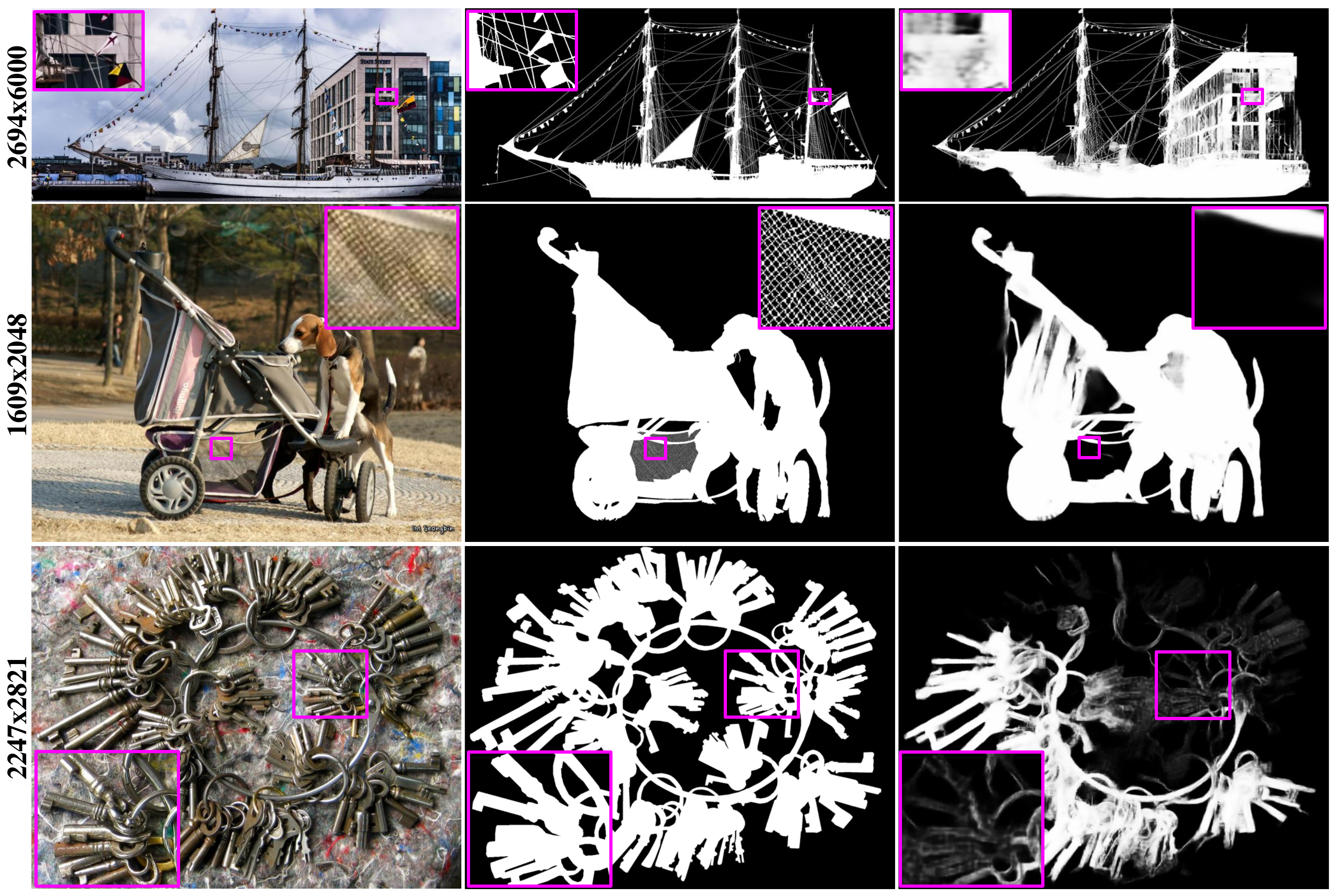}
    \put(16,-1.5) {\small Image}
    \put(52,-1.5) {\small GT}
    \put(82,-1.5) {\small Ours}
    \end{overpic}
    \caption{\small Typical failure cases.}
    \label{fig:fail}
\end{figure*}

\section{Limitations and Future Works} 

\noindent
\textbf{Failure Cases of Our Model.} \figref{fig:fail} shows some typical failure cases of our model. The first row shows the result of a sail ship image. Our model fails in segment two of the masts and the ropes because this region has a cluttered background (a building). The second row shows the segmentation result of a baby carriage. Our model fails in segmenting the mesh-like structure of the carriage since it is too meticulous (just one-pixel width), so that it is hard to be segmented by our model from the input images with the size of $1024\times1024$. The third row illustrates the segmentation result of a key chain with a cluttered background. As can be seen, the color differences between the critical chain and the background are small, which significantly increases the difficulty of the segmentation. 
In summary, the highly accurate dichotomous image segmentation (DIS) is a highly challenging task. There is still a large room for improvement. Therefore, more powerful deep segmentation models are needed to handle larger size input for obtaining very detailed object structures. In contrast, the model size, memory occupation, training, and inference time costs are expected to be affordable on the mainstream GPUs. 

\noindent
\textbf{Limitations of Our DIS5K dataset.} Although our DIS5K is currently the most complex dichotomous segmentation dataset, there is still a large room for improvement. For example, compared with the vast number of categories and the diversified general object classes in the real-world, 225 categories in our DIS5K dataset are far from enough. Therefore, more categories, more samples of specific categories, and more diversified image qualities are needed to further improve the diversity of this dataset. Besides, semi-automatic and highly accurate annotation tools are expected to simplify and boost the ground truth labeling processes. We will explore semi-supervised and weakly supervised methods for further reducing the labeling workloads. In addition, it also requires a set of standard criteria to control the labeling accuracy. 

\noindent
\textbf{Limitations of Our HCE metric.} Our HCE metric provides direct measures of the human correction efforts needed for fixing faulty predictions under certain accuracy requirements. To leverage different accuracy requirements, the erosion \cite{haralick1987image} and dilation \cite{haralick1987image} operations are used to remove small false positive and false negative regions, while the skeleton extraction algorithm \cite{DBLP:journals/cacm/ZhangS84} is used to preserve the structural information of the thin components in the ground truth masks. However, the skeleton extraction algorithm is slow when processing the large-size masks. Therefore, the evaluation of large-scale datasets takes a long time. This issue also happens when computing the weighted F-measure \cite{Margolin2014HowTE}, which uses a distance transform algorithm \cite{DBLP:journals/cvgip/Borgefors86,DBLP:journals/toc/FelzenszwalbH12} to calculate the weights. Therefore, more works need to be conducted on these conventional algorithms, such as skeleton extraction, distance transform, etc., to handle larger and more complicated inputs.

\end{document}